\documentclass[nohyperref]{article}
\usepackage[latin9]{inputenc}
\usepackage{color}
\usepackage{array}
\usepackage{booktabs}
\usepackage{mathtools}
\usepackage{multirow}
\usepackage{amsmath}
\usepackage{amsthm}
\usepackage{amssymb}
\usepackage{graphicx}
\PassOptionsToPackage{normalem}{ulem}
\usepackage{ulem}
\usepackage{microtype}
\usepackage[unicode=true,
 bookmarks=false,
 breaklinks=false,pdfborder={0 0 1},backref=false,colorlinks=false]
 {hyperref}

\makeatletter

\providecommand{\tabularnewline}{\\}

\theoremstyle{plain}
\newtheorem{thm}{\protect\theoremname}
\theoremstyle{definition}
\newtheorem{example}[thm]{\protect\examplename}
\theoremstyle{plain}
\newtheorem{prop}[thm]{\protect\propositionname}

\usepackage[preprint]{neurips_2022}
\allowdisplaybreaks[4]
\usepackage{color}
\usepackage{xcolor}
\usepackage{minitoc}
\setcitestyle{square,numbers,comma}

\makeatother

\providecommand{\examplename}{Example}
\providecommand{\propositionname}{Proposition}
\providecommand{\theoremname}{Theorem}

\begin{document}
\author{%
Zepeng Zhang \\
ShanghaiTech University\\
\texttt{zhangzp1@shanghaitech.edu.cn} \\
\And   
Ziping Zhao \\   
ShanghaiTech University \\
\texttt{zhaoziping@shanghaitech.edu.cn}
}
\title{Towards Understanding Graph Neural Networks:\\
An Algorithm Unrolling Perspective}
\maketitle
\begin{abstract}
The graph neural network (GNN) has demonstrated its superior performance
in various applications. The working mechanism behind it, however,
remains mysterious. GNN models are designed to learn effective representations
for graph-structured data, which intrinsically coincides with the
principle of graph signal denoising (GSD). Algorithm unrolling, a
``learning to optimize'' technique, has gained increasing attention
due to its prospects in building efficient and interpretable neural
network architectures. In this paper, we introduce a class of unrolled
 networks built based on truncated optimization algorithms (e.g.,
gradient descent and proximal gradient descent) for GSD problems.
They are shown to be tightly connected to many popular GNN models
in that the forward propagations in these GNNs are in fact unrolled
networks serving specific GSDs. Besides, the training process of a
GNN model can be seen as solving a bilevel optimization problem with
a GSD problem at the lower level. Such a connection brings a fresh
view of GNNs, as we could try to understand their practical capabilities
from their GSD counterparts, and it can also motivate designing new
GNN models. Based on the algorithm unrolling perspective, an expressive
model named UGDGNN, i.e., unrolled gradient descent GNN, is further
proposed which inherits appealing theoretical properties. Extensive
numerical simulations on seven benchmark datasets demonstrate that
UGDGNN can achieve superior or competitive performance over the state-of-the-art
models. 

\end{abstract}

\section{Introduction}

As the graph-structured data becomes largely available in various
applications \cite{newman2018networks,tolstaya2020learning}, analyses
of high-dimensional data over graph and/or other irregular domains
have recently drawn unprecedented attention from researchers in machine
learning, signal processing, statistics, etc. Among many other approaches,
the graph neural network (GNN) models \cite{kipf2017semi,zhou2020graph,wu2020comprehensive,zhang2020deep,GNNBook2022}
have become prevalent due to their convincing performance and have
been widely used in recommender system \cite{ying2018graph}, computer
vision \cite{satorras2018few}, mathematical optimization \cite{cappart2021combinatorial},
and many other application fields \cite{gilmer2017neural,you2018graph,rong2020self,wu2021graph}.

The GNN models are deep learning-based methods operating on the graph
domain that are designed to extract useful high-dimensional representations
from the raw input graph-structured data (or signals). Typically,
a GNN is a neural network consists of consecutive propagation layers,
each of which contains two steps, namely, the feature aggregation
(FA) step (i.e., each node aggregates the features from its neighborhood)
and the feature transformation (FT) step (i.e., the aggregated features
are further transformed at each node) \cite{hamilton2017inductive,velivckovic2018graph,you2019position,pei2020geom}.
It has been shown that a single FA step in graph convolutional network
(GCN) \cite{kipf2017semi} and simple graph convolution network (SGC)
\cite{wu2019simplifying} can be seen as a special form of Laplacian
smoothing \cite{li2018deeper,hoang2021revisiting,yang2021rethinking}.
Inspired by this result, some recent works aim at interpreting one
FA step in some other GNNs with a Laplacian regularized optimization
problem \cite{zhu2021interpreting,ma2021unified}. However, all the
existing analyses in the literature isolate the consecutive propagation
layers of the GNN models and further merely focus on the FA step in
one propagation layer. Hence, their results neglect not only the accompanying
indispensable FT step in each layer but also the propagation nature
between consecutive layers of GNNs. Failing to capture the GNN structures
as a whole, such over-simplified analyses will inevitably lead to
conclusions that are discrepant with the practical capabilities of
GNNs \cite{yang2020revisiting,zhou2021understanding,cong2021provable}.
Then a natural research question arises: \emph{Can we interpret the
existing GNN models with multiple consecutive feature propagation
layers in a holistic way? }

This paper gives an affirmative answer to the above question from
the perspective of algorithm unrolling \cite{gregor2010learning}.
Algorithm unrolling falls into the ``learning to optimize'' paradigm
and is regarded to be a promising technique for efficient and interpretable
neural network architecture design \cite{gregor2010learning,monga2021algorithm,chen2021learning}.
The idea of algorithm unrolling is to transform a truncated iterative
algorithm (i.e., an algorithm with finite steps) into a neural network,
called an ``unrolled network,'' by converting some parameters of the
updating steps in the original iterative algorithms into learnable
weights in the unrolled network. 

In this paper, we initiate our discussion with a bilevel optimization
problem \cite{dempe2020bilevel}, where the upper-level problem is
to minimize a task-tailored supervised loss and in the lower level
is a graph signal denoising (GSD) optimization problem \cite{shuman2013emerging,stankovic2019introduction}.
The GSD problems aim at recovering the denoised graph signals from
certain noisy input graph signals, which conceptually share the same
spirit as the GNN models, i.e., learning effective graph representations
from the raw input graph features. In practice, formulations for GSD
problems are diverse due to the varying intended denoising targets.
Analogously, different GNN models exhibit different model structures
and/or parameters. After that, we present two unrolled networks for
solving the GSD problems, namely, unrolled gradient descent (GD) network
and unrolled proximal gradient descent (ProxGD) network. It turns
out that a class of widely used GNN models are specializations or
variants of these two unrolled networks. Therefore, training a GNN
via backpropagation can be regarded as solving a bilevel optimization
problem. On this basis, some novel inherent characteristics of the
existing GNN models can further be revealed from their GSD counterparts
and new GNN models with desired properties can also be developed. 

To make it clear, main contributions of this paper are summarized
in the following. 
\begin{itemize}
\item We present a unified algorithm unrolling perspective on GNNs, based
on which we demonstrate that many existing GNN models can be naturally
interpreted as either an unrolled GD network or an unrolled ProxGD
network for solving certain GSD problems. With this, for the first
time, we accomplish a holistic understanding of GNNs as solving GSD
problems. 
\item The fresh algorithm unrolling perspective on GNNs implies that training
a GNN model can be naturally seen as solving a bilevel optimization
problem with supervised loss at the upper level and a GSD problem
at the lower level, which is tackled via an interpretable unrolled
neural network corresponding to the consecutive propagation layers
in the GNN model.
\item Based on the algorithm unrolling perspective, a novel GNN with appealing
expressive power is proposed leveraging the unrolled GD network for
GSDs, hence named as UGDGNN. 
\item Extensive experimental results on seven benchmark datasets verify
the effectiveness of UGDGNN and some of the theoretical conjectures
made along this paper. The applaudable empirical results signify the
promising prospects of interpreting existing GNNs as well as designing
new GNN architectures from the algorithm unrolling perspective.
\end{itemize}
Due to space limit, proofs for the main results and further discussions
are all deferred to the Appendix.

\section{Preliminaries}

\textbf{Notations.} An undirected unweighted graph with self-loops
is denoted as $\mathcal{G}=(\mathcal{V},\mathcal{E})$, where $\mathcal{V}=\{1,\ldots,|\mathcal{V}|\}$
and $\mathcal{E}$ are the node set and the edge set, respectively.
Suppose the $n$-th ($n\in\mathcal{V}$) node is associated with a
feature vector $\mathbf{x}_{n}$, $\mathbf{X}=[\mathbf{x}_{1},\ldots,\mathbf{x}_{|\mathcal{V}|}]^{T}$
represents the overall graph feature matrix. Each column of $\mathbf{X}$
is a one-dimensional graph signal and $\mathbf{X}$ is essentially
a multi-dimensional graph signal. The graph adjacency structure can
be encoded by the adjacency matrix $\mathbf{A}\in\mathbb{R}^{|\mathcal{V}|\times|\mathcal{V}|}$.
The degree matrix $\mathbf{D}\in\mathbb{R}^{|\mathcal{V}|\times|\mathcal{V}|}$
is defined by $\mathbf{D}=\mathsf{diag}(\mathbf{A}\mathbf{1})$. The
Laplacian matrix is $\mathbf{L}=\mathbf{D}-\mathbf{A}$ and we also
have $\mathbf{L}=\mathbf{B}^{T}\mathbf{B}$ with $\mathbf{B}$ being
an oriented edge-node incident matrix (the orientation can be arbitrary).
In the GNN literature, the symmetric normalized adjacency matrix $\hat{\mathbf{A}}=\mathbf{D}^{-\frac{1}{2}}\mathbf{A}\mathbf{D}^{-\frac{1}{2}}$
and the symmetric normalized Laplacian matrix $\hat{\mathbf{L}}=\mathbf{I}-\hat{\mathbf{A}}=\mathbf{D}^{-\frac{1}{2}}\mathbf{L}\mathbf{D}^{-\frac{1}{2}}=\hat{\mathbf{B}}^{T}\hat{\mathbf{B}}$
with $\hat{\mathbf{B}}=\mathbf{B}\mathbf{D}^{-\frac{1}{2}}$ are more
commonly used. With a slight abuse of notation, in the following sections
we will adopt the unhatted notations to represent the normalized graph
matrices for notational simplicity. 

\subsection{Graph Signal Denoising Problems}

As a fundamental problem in signal processing over graphs, GSD aims
to recover a denoised graph signal $\mathbf{H}$ from the noisy graph
signal $\mathbf{X}$ \cite{shuman2013emerging,kalofolias2016learn,stankovic2019introduction}.
Under the smoothness assumption (i.e., signals in adjacent nodes are
close in value), a vanilla GSD optimization problem is given as follows:
\begin{equation}
\begin{aligned} & \underset{\mathbf{H}\in\mathcal{H}}{\mathrm{min}} &  & \alpha\left\Vert \mathbf{H}-\mathbf{X}\right\Vert _{F}^{2}+\beta\mathrm{tr}\bigl(\mathbf{H}^{T}\mathbf{L}\mathbf{H}\bigr),\end{aligned}
\label{Formulation: GSD}
\end{equation}
where $\mathcal{H}$ represents some possible constraints on $\mathbf{H}$,
and $\alpha$, $\beta\geq0$ adjust the relative importance of the
two terms in the objective. To be specific, the first term is the
signal fidelity term that guides $\mathbf{H}$ to be as close to $\mathbf{X}$
as possible, and the second term is the signal smoothing term (actually
a Laplacian smoothing term, a.k.a. Laplacian regularization) that
promotes smoothness of the denoised graph signal $\mathbf{H}$ (note
that since $\mathbf{L}=\mathbf{B}^{T}\mathbf{B}$ we have $\mathrm{tr}(\mathbf{H}^{T}\mathbf{L}\mathbf{H})\negthinspace=\negthinspace||\mathbf{B}\mathbf{H}||_{F}^{2}$).
As in the widely studied generalized regression and generalized low-rank
approximation problems \cite{srebro2003weighted}, we can always replace
the Frobenius norms in \eqref{Formulation: GSD} with their weighted
alternatives defined by $||\mathbf{M}||_{\mathbf{T}}^{2}\coloneqq\mathrm{tr}(\mathbf{M}\mathbf{T}\mathbf{M}^{T})$
where $\mathbf{T}$ is symmetric. Then, we obtain the following
general GSD problem:
\begin{equation}
\begin{aligned} & \underset{\mathbf{H}}{\mathrm{min}} &  & \underset{\eqqcolon L\left(\mathbf{H};\boldsymbol{\Theta},\mathbf{X},\mathbf{B}\right)}{\underbrace{\alpha\bigl\Vert\mathbf{H}-\mathbf{X}\bigr\Vert_{\mathbf{T}_{\alpha}}^{2}+\beta\bigl\Vert\mathbf{B}\mathbf{H}\bigr\Vert_{\mathbf{T}_{\beta}}^{2}+r(\mathbf{H})}},\end{aligned}
\label{Formulation: GSD reg}
\end{equation}
where $\mathbf{T}_{\alpha}$ and $\mathbf{T}_{\beta}$ generally represent
certain prior information on the target of GSD (For instance, $\mathbf{T}_{\alpha}$
could be chosen as the inverse covariance matrix of the noise, and
$\mathbf{T}_{\beta}$ can be used to enforce different levels of smoothness
on different nodes. Specially, a positive $t_{\beta,ii}$ promotes
smoothness while a negative $t_{\beta,ii}$ promotes non-smoothness
between nodes connected by the $i$-th edge, which facilitates learning
over both homogeneous and heterogeneous graphs.) and $r(\mathbf{H})$
is a regularization term which can be used to impose constraints or
other penalizations. We have denoted the objective function as $L\left(\mathbf{H};\boldsymbol{\Theta},\mathbf{X},\mathbf{B}\right)$
where $\boldsymbol{\Theta}$ collects all the model parameters including
$\alpha$, $\beta$, $\mathbf{T}_{\alpha}$, and $\mathbf{T}_{\beta}$.
In addition, problem \eqref{Formulation: GSD reg} will degenerate
to the vanilla GSD problem \eqref{Formulation: GSD} when $\mathbf{T}_{\alpha}=\mathbf{T}_{\beta}=\mathbf{I}$
and $r(\mathbf{H})=I_{\mathcal{H}}(\mathbf{H})$, where $I(\cdot)$
denotes the indicator function taking value 0 if $\mathbf{H}\in\mathcal{H}$
and $+\infty$ otherwise. Based on the GSD problems defined above,
we can see the goals of GSD are intrinsically similar to those of
GNN, i.e., extracting certain ``useful information'' from the noisy
inputs.

\subsection{Algorithm Unrolling}

Algorithm unrolling, a.k.a. algorithm unfolding, is a ``learning to
optimize'' paradigm for neural network architecture design, which
``unrolls'' a truncated classical iterative algorithm (i.e., an iterative
algorithm only with finite iteration steps) into a neural network
\cite{gregor2010learning,monga2021algorithm,chen2021learning}, named
an ``unrolled network.'' To design an unrolled network, some of the
parameters in one updating step from the original iterative algorithm
will be translated to be the trainable weights in one network layer,
and a finite number of layers are then concatenated together. Such
a trained neural network will naturally carry out the interpretation
of a ``parameter-optimized iterative algorithm'' \cite{yang2021implicit}.
Therefore, passing through an unrolled network with a finite number
of layers amounts to execute an iterative algorithm for a finite number
of steps, which brings interpretability to the neural network.

In constructing an unrolled network, a key procedure is to determine
how to parameterize the underlying iterative algorithm (considerations
may include which parameters are translated to be the learnable weights
in the network, setting the parameters in different layers to be shared
or independent, etc.). The algorithm unrolling scheme was first used
to unroll the famous iterative soft-thresholding algorithm \cite{gregor2010learning},
i.e., proximal gradient, for sparse coding problems, where hyperparameters
in the algorithm are chosen to be learnable and set to be shared over
different layers, i.e., a recurrent structure. After that, some other
works \cite{hershey2014deep,borgerding2016onsager,chen2018theoretical,liu2019alista}
drop the recurrent structure scheme and untie the learnable weights;
i.e., weights in different layers can be different and hence optimized
independently. These models with untying parameters can enlarge the
capacity of the unrolled networks \cite{hershey2014deep}. Although
research on the theoretical characterizations between these unrolled
networks and their analytic iterative algorithm counterparts are still
evolving \cite{monga2021algorithm,chen2021learning}, interpreting
popular neural network architectures from the algorithm unrolling
perspective can always enrich the understanding of their capacities.
For example, authors in \cite{papyan2017convolutional} analyzed the
properties of the convolutional neural networks by viewing them as
unrolled networks. To the best of our knowledge, the relationship
between algorithm unrolling and the existing GNN models remains unexplored.

\section{A Unified Algorithm Unrolling Perspective on GNN Models}

We launch our discussion on the ``intimate correspondence'' between
GNN models and GSD optimization problems through investigating a bilevel
optimization problem \cite{dempe2020bilevel} defined as follows:
\begin{equation}
\begin{aligned} & \underset{\boldsymbol{\theta}\in\varTheta,\bar{\mathbf{X}}}{\mathrm{min}} &  & U\left(\mathbf{Y},p_{\mathsf{pos}}(\bar{\mathbf{X}})\right) & \mathsf{(upper)}\\
 & \;\:\mathrm{s.t.} &  & \bar{\mathbf{X}}\in\mathrm{arg}\min_{\mathbf{H}}L\Biggl(\mathbf{H};\boldsymbol{\Theta},\underset{\eqqcolon\mathbf{X}}{\underbrace{p_{\mathsf{pre}}\left(\tilde{\mathbf{X}}\right)}},\mathbf{B}\Biggr), & \mathsf{(lower)}
\end{aligned}
\label{Eq: Bilevel Formulation}
\end{equation}
where the optimization variables are all the model parameters $\boldsymbol{\theta}\coloneqq\{\boldsymbol{\Theta},\text{prameters in }p_{\mathsf{pos}}\text{\text{ and }}p_{\mathsf{pre}}\}$
with $\varTheta$ denoting the parameter space and the output $\bar{\mathbf{X}}$
of the lower-level problem. The objective $U$ in the upper level
represents the supervised loss for downstream tasks like clustering,
classification, ranking, etc., with $\mathbf{Y}$ denoting the label
matrix and $p_{\mathsf{pos}}(\bar{\mathbf{X}})$ representing some
possible post-processing procedures for $\bar{\mathbf{X}}$. The lower-level
problem is a GSD problem as defined in \eqref{Formulation: GSD reg}.
Instead of denoising the raw input graph signal $\tilde{\mathbf{X}}$
directly, certain pre-processing operations may also be imposed before
it is fed into the GSD problem, which is generally denoted by $p_{\mathsf{pre}}(\tilde{\mathbf{X}})$,
whose output is defined as $\mathbf{X}$. 
\begin{example}[Semi-supervised node classification]
Consider a semi-supervised node classification problem, where we
can only access labels of a subset of nodes, i.e., $\mathcal{V}^{\prime}\subset\mathcal{V}$.
The target is to train a GNN model based on the raw node features
of all nodes $\tilde{\mathbf{X}}$ and the node labels in the subset
$\{\mathbf{y}_{i}\}_{i\in\mathcal{V}^{\prime}}$, and to expect it
can generalize well on the unlabeled node set $\mathcal{V}\setminus\mathcal{V}^{\prime}$.
In this case, we can choose $U$ as the cross-entropy loss function,
$p_{\mathsf{pos}}$ as the $\mathrm{softmax}$ function, and $p_{\mathsf{pre}}$
as a multi-layer perceptron.
\end{example}

Generally speaking, solving the nested optimization problem \eqref{Eq: Bilevel Formulation}
is challenging as it involves solving the upper- and lower-level problems
simultaneously. If an analytical solution to the lower-level problem
can be obtained and then substituted into the upper level, the bilevel
problem will reduce to a single-level one, which could be easier to
handle. In case that analytical solutions for the lower-level problem
are not available (or expensive to compute), one can resort to an
iterative algorithm (e.g., GD, ProxGD, Newton's methods, etc.) to
generate an approximate solution by conducting a finite number of
iterations. To expand the capacity of an iterative algorithm, the
algorithm unrolling technique can be used to construct an unrolled
network. Given an initialization $\mathbf{H}^{(0)}$, passing it through
an $K$-layer unrolled network imitates the behavior of executing
an iterative algorithm for $K$ steps, generating the sequence $\{\mathbf{H}^{(k)}\}_{k=1}^{K}$.
Under certain conditions, convergence of $\{\mathbf{H}^{(k)}\}_{k=1}^{K}$
to $\bar{\mathbf{X}}$ can be established \cite{chen2018theoretical}.
With a slight abuse of notation, $\bar{\mathbf{X}}$ and $\mathbf{H}^{(K)}$
will be used interchangeably in discussing GNNs.

Specifically, applying the idea of algorithm unrolling to an iterative
algorithm that solves the GSD optimization problem at the lower level
of \eqref{Eq: Bilevel Formulation}, an unrolled network for graph
representation leaning will be obtained. Its parameters are to be
learned through optimizing the loss at the upper level. In the following,
we show that a class of widely used existing GNN models naturally
emerge out of unrolled GD or unrolled ProxGD networks for solving
GSD problems, based on which training these GNNs is equivalent to
solving a bilevel optimization problem with certain GSDs at the lower
level.

\subsection{GNNs as Unrolled GD Networks for GSDs \label{Section: Unrolling GD}}

It turns out that many popular GNN models can be seen as unrolled
GD networks for GSDs. Denote the objective in \eqref{Formulation: GSD reg}
as $L(\mathbf{H})$ for brevity. The update in GD with properly chosen
stepsize $\eta^{(k)}>0$ is
\[
\mathbf{H}^{(k)}=\mathbf{H}^{(k-1)}-\eta^{(k)}\nabla L(\mathbf{H}^{(k-1)}).
\]
With $r(\mathbf{H})=0$, the GSD objective $L(\mathbf{H})$ in \eqref{Formulation: GSD reg}
is smooth and the gradient of $L(\mathbf{H})$ is
\begin{equation}
\nabla L(\mathbf{H})=2\alpha\bigl(\mathbf{H}-\mathbf{X}\bigr)\mathbf{T}_{\alpha}+2\beta\mathbf{B}^{T}\mathbf{B}\mathbf{H}\mathbf{T}_{\beta},\label{eq:GD exp}
\end{equation}
and accordingly the GD step is given by
\[
\mathbf{H}^{(k)}=\mathbf{H}^{(k-1)}\bigl(\mathbf{I}-2\eta^{(k)}\alpha\mathbf{T}_{\alpha}-2\eta^{(k)}\beta\mathbf{T}_{\beta}\bigr)+2\eta^{(k)}\beta\mathbf{A}\mathbf{H}^{(k-1)}\mathbf{T}_{\beta}+2\eta^{(k)}\alpha\mathbf{X}\mathbf{T}_{\alpha},
\]
where we have used the relation $\mathbf{B}^{T}\mathbf{B}=\mathbf{I}-\mathbf{A}$.
Following the algorithm unrolling paradigm, by letting the stepsize
parameter $\eta^{(k)}$ and model parameters $\boldsymbol{\Theta}$
(i.e., $\alpha$, $\beta$, $\mathbf{T}_{\alpha}$, and $\mathbf{T}_{\beta}$)
in the GSD problems to be learnable and concatenating $K$ unrolled
GD layers together, a GNN model can be naturally constructed as an
unrolled GD neural network. 
\begin{thm}
\label{Thm:Unrolling GD} A general $K$-layer GNN model can be obtained
by unrolling $K$ GD steps for an unconstrained GSD problem \eqref{Formulation: GSD reg}
(i.e., $r(\mathbf{H})=0$), in which the $k$-th propagation layer
is given by
\begin{equation}
\begin{aligned} & \mathbf{H}^{(k)}=\mathbf{H}^{(k-1)}\bigl(\mathbf{I}-2\eta^{(k)}\alpha^{(k)}\mathbf{T}_{\alpha}^{(k)}-2\eta^{(k)}\beta^{(k)}\mathbf{T}_{\beta}^{(k)}\bigr)\\
 & \hspace{2.55cm}+2\eta^{(k)}\beta^{(k)}\mathbf{A}\mathbf{H}^{(k-1)}\mathbf{T}_{\beta}^{(k)}+2\eta^{(k)}\alpha^{(k)}\mathbf{X}\mathbf{T}_{\alpha}^{(k)},\ \text{for}\ k=1,\ldots,K,
\end{aligned}
\label{Unrolling Step: GD}
\end{equation}
where $\eta^{(k)}$, $\alpha^{(k)}$, $\beta^{(k)}$, $\mathbf{T}_{\alpha}^{(k)}$,
and $\mathbf{T}_{\beta}^{(k)}$ are the learnable parameters in the
$k$-th network layer. 
\end{thm}

In an unrolled GD layer \eqref{Unrolling Step: GD}, i.e., a propagation
layer (including one FA step and one FT step) of a general GNN model,
the parameters $\eta^{(k)}$, $\alpha^{(k)}$, and $\beta^{(k)}$
jointly control the strength of neighborhood aggregation, residual
connection, and initial residual connection. Unlike the existing works
\cite{pan2020unified,zhu2021interpreting,ma2021unified} that only
interpret the FA step in one propagation layer without taking the
indispensable transformation weight matrices into account and hence
may be myopic or hyperopic, our interpretation of a propagation layer
as given in \eqref{Unrolling Step: GD} can be easily extended to
interpret the consecutive propagation nature of GNN models in a holistic
manner. It should also be noted that although the unrolled GD layer
in \eqref{Unrolling Step: GD} seems to be very comprehensive, this
neural network is actually a structured and interpretable model because
it is unrolled based on GD for a prescribed GSD problem. It can be
shown that existing GNNs can be seen as specializations or variants
of this unrolled GD network.

In the following, we will elaborate four representative GNNs as unrolled
GD networks for GSDs, that is, SGC \cite{wu2019simplifying}, approximate
personalized propagation of neural predictions (APPNP) \cite{klicpera2019predict},
jumping knowledge networks (JKNet) \cite{xu2018representation}, and
generalized PageRank GNN (GPRGNN) \cite{chien2021adaptive}. Note
that for all models, the specific parameterization schemes in unrolled
networks are elaborated in the Appendix. 

\textbf{\textcolor[RGB]{128,42,42}{SGC.}} The SGC \cite{wu2019simplifying}
performs a finite number of FA steps (i.e., recursively premultiplying
$\mathbf{A}$ to the current feature) followed by a linear transformation.
To be specific, a $K$-layer SGC is defined as
\begin{equation}
\bar{\mathbf{X}}=\mathbf{A}^{K}\mathbf{X}\mathbf{W},\label{eq:SGC prop}
\end{equation}
where $\mathbf{W}$ is a learnable parameter. The relation between
SGC and GSD is given in Proposition \ref{thm:SGC}.
\begin{prop}
\label{thm:SGC} Passing through a $K$-layer SGC model is equivalent
to solving a GSD problem \eqref{Formulation: GSD reg} with $\alpha=0$
and $r(\mathbf{H})=\beta||\mathbf{H}||_{\mathbf{I}-\mathbf{T}_{\beta}}^{2}$
via a $K$-layer unrolled GD network. 
\end{prop}

In Proposition \ref{thm:SGC}, we have assumed the learnable weight
matrix $\mathbf{W}$ is induced from parameter $\boldsymbol{\Theta}$,
while it actually can also be cast as a part of $p_{\mathsf{pre}}$
or $p_{\mathsf{pos}}$ (refer to Appendix for detailed discussions). 

\textbf{\textcolor[RGB]{128,42,42}{PPNP \& APPNP.}} It has been discussed
in \cite{klicpera2019predict} that as $K\rightarrow+\infty$, the
aggregation scheme in SGC will lead to a limit point that has a similar
expression as in PageRank \cite{page1998pagerank}. Based on the relationship
between SGC and PageRank, the PPNP model \cite{klicpera2019predict}
was proposed based on personalized PageRank \cite{page1998pagerank},
whose propagation mechanism is
\[
\bar{\mathbf{X}}=\gamma\bigl(\mathbf{I}-(1-\gamma)\mathbf{A}\bigr)^{-1}\mathbf{X},
\]
where $\gamma$ is a given teleport probability. To avoid the computationally
inefficient matrix inverse operation, an approximation of PPNP called
APPNP \cite{klicpera2019predict} was proposed. A $K$-layer APPNP
is
\[
\bar{\mathbf{X}}=\mathbf{H}^{(K)},\ \ \ \mathbf{H}^{(k)}=\left(1-\gamma\right)\mathbf{A}\mathbf{H}^{(k-1)}+\gamma\mathbf{X},\ \text{for}\ k=1,\ldots,K.
\]
The relation between PPNP \& APPNP and GSD problems is summarized
in the following proposition.
\begin{prop}
\label{thm:APPNP} The PPNP model analytically solves the GSD problem
\eqref{Formulation: GSD reg} with $\beta=1-\alpha$, $\mathbf{T}_{\alpha}=\mathbf{T}_{\beta}=\mathbf{I}$,
and $r(\mathbf{H})=0$, while a $K$ layer APPNP model solves it via
a $K$-layer unrolled GD network. Moreover, the teleport probability
$\gamma$ in PPNP \& APPNP is equal to $\alpha$ in the underlying
GSD problem.
\end{prop}

Based on Proposition \ref{thm:APPNP}, a smaller $\gamma$ in APPNP
\& PPNP, i.e., a smaller $\alpha$ and hence a relatively larger $\beta$
in GSD, can enforce the graph signal to be smoother, corroborating
the meaning of parameter $\gamma$.

\textbf{\textcolor[RGB]{128,42,42}{JKNet.}} In SGC, the nodes in
graph can only access information from a fixed-size neighborhood.
To make each node adaptively leverage the information from its effective
neighborhood size, the JKNet model \cite{xu2018representation} selectively
combines the intermediate node representations from each layer into
its output. The JKNet model permits general representation combination
schemes, but for simplicity we consider the summation scheme here.
A $K$-layer JKNet applies the following operation to the input $\mathbf{X}$:
\begin{equation}
\bar{\mathbf{X}}=\sum_{k=0}^{K}\mathbf{A}^{k}\mathbf{X}\mathbf{W}^{(k)},\label{eq:JKNet}
\end{equation}
where $\{\mathbf{W}^{(k)}\}_{k=1}^{K}$ are the learnable weights.
The relation between JKNet and GSD is given below.
\begin{prop}
\label{thm: JKNet} By concatenating $K$ unrolled GD layers for solving
a GSD problem \eqref{Formulation: GSD reg} with $\alpha=\beta$,
$\mathbf{T}_{\alpha}=\mathbf{I}-\mathbf{T}_{\beta}$, and $r(\mathbf{H})=0$,
we obtain a $K$-layer JKNet model.  
\end{prop}

\textbf{\textcolor[RGB]{128,42,42}{GPRGNN.}} The generalized PageRank
methods were first used in unsupervised graph clustering and demonstrated
significant performance improvements over personalized PageRank methods
\cite{kloumann2017block,li2019optimizing}. Inspired by that, the
GPRGNN model \cite{chien2021adaptive} was brought up based on the
generalized PageRank technique, which associates each step of feature
aggregation with a learnable weight. A $K$-layer GPRGNN model is
given as follows:
\begin{equation}
\bar{\mathbf{X}}=\sum_{k=0}^{K}\gamma^{(k)}\mathbf{A}^{k}\mathbf{X},\label{eq:GPRGNN}
\end{equation}
where $\{\gamma^{(k)}\}_{k=1}^{K}$ are learnable weights. The relation
between GPRGNN and GSD is provided below.
\begin{prop}
\label{thm: GPRGNN} A $K$-layer GPRGNN is equivalent to a $K$-layer
unrolled GD network for solving a GSD problem \eqref{Formulation: GSD reg}
with $\beta=1-\alpha$, $\mathbf{T}_{\alpha}=\mathbf{T}_{\beta}=\mathbf{I}$,
and $r(\mathbf{H})=0$.
\end{prop}

Based on Proposition \ref{thm:APPNP} and Proposition \ref{thm: GPRGNN},
we can conclude that APPNP and GPRGNN solve the same underlying GSD
problem. Besides, GPRGNN can be seen as a generalization of APPNP
by untying the shared parameter $\gamma$ in different unrolled GD
layers. Therefore, GPRGNN is supposed to be more expressive and hence
is expected to attain better performance than APPNP.

\subsection{GNNs as Unrolled ProxGD Networks for GSDs \label{Section: ProxGD}}

In the previous section, we have considered several GNNs only with
linear FT steps. However, there are many GNN models that apply nonlinear
FT steps. Considering the fact that every standard activation function
(e.g., $\mathrm{ReLU}$, $\mathrm{ELU}$, and leaky $\mathrm{ReLU}$)
can be derived from a proximity operator \cite{combettes2020deep},
GNNs with nonlinear FTs can be naturally interpreted as unrolled ProxGD
networks for GSDs. 

\begin{table}
\caption{The GSD objectives for different GNN models from the algorithm unrolling
perspective.\label{tab:GSD-GNN}}
\centering{}\resizebox{1.0 \textwidth}{!}{%
\begin{tabular}{cll}
\toprule 
\textbf{Models} & \textbf{$K$-layer GNN Models} & \textbf{GSD Objectives}\tabularnewline
\midrule
\midrule 
SGC & $\bar{\mathbf{X}}=\mathbf{A}^{K}\mathbf{X}\mathbf{W}$ & $\bigl\Vert\mathbf{B}\mathbf{H}\bigr\Vert_{\mathbf{T}_{\beta}}^{2}+\bigl\Vert\mathbf{H}\bigr\Vert_{\mathbf{I}-\mathbf{T}_{\beta}}^{2}$\tabularnewline
\midrule 
PPNP & $\bar{\mathbf{X}}=\gamma\bigl(\mathbf{I}-(1-\gamma)\mathbf{A}\bigr)^{-1}\mathbf{X}$ & \multirow{2}{*}{$\alpha\bigl\Vert\mathbf{H}-\mathbf{X}\bigr\Vert_{F}^{2}+(1-\alpha)\bigl\Vert\mathbf{B}\mathbf{H}\bigr\Vert_{F}^{2}$}\tabularnewline
\cmidrule{1-2} \cmidrule{2-2} 
APPNP & $\bar{\mathbf{X}}=\mathbf{H}^{(K)},\ \mathbf{H}^{(k)}=(1-\gamma)\mathbf{A}\mathbf{H}^{(k-1)}+\gamma\mathbf{X},\ \ k=1,\ldots,K$ & \tabularnewline
\midrule 
JKNet & $\bar{\mathbf{X}}=\sum_{k=0}^{K}\mathbf{A}^{k}\mathbf{X}\mathbf{W}^{(k)}$ & $\bigl\Vert\mathbf{H}-\mathbf{X}\bigr\Vert_{\mathbf{T}_{\alpha}}^{2}+\bigl\Vert\mathbf{B}\mathbf{H}\bigr\Vert_{\mathbf{I}-\mathbf{T}_{\alpha}}^{2}$\tabularnewline
\midrule 
GPRGNN & $\bar{\mathbf{X}}=\sum_{k=0}^{K}\gamma^{(k)}\mathbf{A}^{k}\mathbf{X}$ & $\alpha\bigl\Vert\mathbf{H}-\mathbf{X}\bigr\Vert_{F}^{2}+(1-\alpha)\bigl\Vert\mathbf{B}\mathbf{H}\bigr\Vert_{F}^{2}$\tabularnewline
\midrule 
GCN & $\bar{\mathbf{X}}=\mathbf{H}^{(K)},\ \mathbf{H}^{(k)}=\mathrm{ReLU}\left(\mathbf{A}\mathbf{H}^{(k-1)}\mathbf{W}^{(k)}\right),\ \ k=1,\ldots,K$ & $\bigl\Vert\mathbf{B}\mathbf{H}\bigr\Vert_{\mathbf{T}_{\beta}}^{2}+\bigl\Vert\mathbf{H}\bigr\Vert_{\mathbf{I}-\mathbf{T}_{\beta}}^{2}+I_{\{h_{ij}\geq0,\ \forall i,j\}}(\mathbf{H})$\tabularnewline
\midrule 
\multirow{2}{*}{GCNII} & $\bar{\mathbf{X}}=\mathbf{H}^{(K)},\ \mathbf{H}^{(k)}=\mathrm{ReLU}\Bigl(\bigl(\zeta\mathbf{A}\mathbf{H}^{(k-1)}+(1-\zeta)\mathbf{X}\bigr)$ & $\left(1-\beta\right)\bigl\Vert\mathbf{H}-\mathbf{X}\bigr\Vert_{\mathbf{T}_{\beta}}^{2}+\beta\bigl\Vert\mathbf{B}\mathbf{H}\bigr\Vert_{\mathbf{T}_{\beta}}^{2}$\tabularnewline
 & $\hspace{3cm}\times\bigl(\xi\mathbf{W}^{(k)}+(1-\xi)\mathbf{I}\bigr)\Bigr),\ \ k=1,\ldots,K$ & $\hspace{1.55cm}+\bigl\Vert\mathbf{H}\bigr\Vert_{\mathbf{I}-\mathbf{T}_{\beta}}^{2}+I_{\{h_{ij}\geq0,\ \forall i,j\}}(\mathbf{H})$\tabularnewline
\midrule 
\multirow{2}{*}{AirGNN} & $\bar{\mathbf{X}}=\mathbf{H}^{(K)},\ \mathbf{h}_{i}^{(k)}=\mathrm{ReLU}\Bigl(1-\frac{1-\gamma}{2\gamma||\mathbf{A}\mathbf{h}_{i}^{(k-1)}-\mathbf{x}_{i}||_{2}}\Bigr)$ & \multirow{2}{*}{$\beta\bigl\Vert\mathbf{B}\mathbf{H}\bigr\Vert_{F}^{2}+(1-\beta)\bigl\Vert\mathbf{H}-\mathbf{X}\bigr\Vert_{2,1}$}\tabularnewline
 & $\hspace{2.1cm}\times\bigl(\mathbf{A}\mathbf{h}_{i}^{(k-1)}-\mathbf{x}_{i}\bigr)+\mathbf{x}_{i},\ i\in\mathcal{V},\ k=1,\ldots,K$ & \tabularnewline
\midrule 
UGDGNN & $\bar{\mathbf{X}}=\sum_{k=0}^{K}\gamma^{(k)}\mathbf{A}^{k}\mathbf{X}\bigl(\zeta^{(k)}\mathbf{I}+\xi^{(k)}\mathbf{W}^{(k)}\bigr)$ & $\alpha\bigl\Vert\mathbf{H}-\mathbf{X}\bigr\Vert_{\mathbf{T}_{\alpha}}^{2}+\beta\bigl\Vert\mathbf{B}\mathbf{H}\bigr\Vert_{\mathbf{T}_{\beta}}^{2}$\tabularnewline
\bottomrule
\end{tabular}}
\end{table}

Consider Problem \eqref{Formulation: GSD reg} with $r(\mathbf{H})=I_{\mathcal{H}}(\mathbf{H})$,
we can apply the ProxGD algorithm for problem resolution and the $k$-th
update step in ProxGD with properly chosen stepsize $\eta^{(k)}>0$
are given as
\begin{align*}
\mathbf{H}^{(k)} & =\mathrm{prox}_{I}\bigl(\mathbf{H}^{(k-1)}-\eta^{(k)}\nabla\tilde{L}(\mathbf{H}^{(k-1)})\bigr),
\end{align*}
where we use $\tilde{L}$ to denote the smooth part in $L$. By applying
the algorithm unrolling idea to the ProxGD algorithm, we have the
following result.
\begin{thm}
\label{Thm:Unrolling ProxGD} By unrolling $K$ ProxGD steps for a
constrained GSD problem \eqref{Formulation: GSD reg}, we obtain a
general $K$-layer GNN model with the $k$-th ($k=1,\ldots,K$) propagation
layer defined as follows:
\begin{equation}
\begin{aligned}\mathbf{H}^{(k)} & =\mathrm{prox}_{I}\bigl(\mathbf{H}^{(k-1)}\bigl(\mathbf{I}-2\eta^{(k)}\alpha^{(k)}\mathbf{T}_{\alpha}^{(k)}-2\eta^{(k)}\beta^{(k)}\mathbf{T}_{\beta}^{(k)}\bigr)\\
 & \hspace{2.55cm}+2\eta^{(k)}\beta^{(k)}\mathbf{A}\mathbf{H}^{(k-1)}\mathbf{T}_{\beta}^{(k)}+2\eta^{(k)}\alpha^{(k)}\mathbf{X}\mathbf{T}_{\alpha}^{(k)}\bigr),\ \text{for}\ k=1,\ldots,K,
\end{aligned}
\label{Unrolling Step: PGD}
\end{equation}
where $\eta^{(k)}$, $\alpha^{(k)}$, $\beta^{(k)}$, $\mathbf{T}_{\alpha}^{(k)}$,
and $\mathbf{T}_{\beta}^{(k)}$ are the learnable parameters in the
$k$-th network layer. 
\end{thm}

Specifically, suppose we have the prior information that the graph
signals of interest are non-negative, i.e., a non-negative GSD problem,
we can set a constraint as $\mathcal{H}=\{h_{ij}\geq0,\ \forall i,j\}$.
Due to the fact that $\mathrm{prox}_{I(\{x\geq0\})}(x)=\mathrm{ReLU}(x)$
\cite{combettes2020deep}, the activation function $\mathrm{ReLU}$
in a class of GNNs could be interpreted as unrolled ProxGD networks
for non-negative GSDs. Similar to unrolled GD network, this model
is highly structured and encompasses some existing GNN models. In
the following, three representative GNNs that perform nonlinear FT
steps are showcased, namely, GCN \cite{kipf2017semi}, GCN with initial
residual and identity mapping (GCNII) \cite{chen2020simple}, and
GNN with adaptive residual (AirGNN) \cite{liu2021graph}.

\textbf{\textcolor[RGB]{128,42,42}{GCN.}} Although appearing earlier
than SGC in the literature, the GCN model \cite{kipf2017semi} can
be regarded as a nonlinear extension of SGC by adding the nonlinear
activation functions $\mathrm{ReLU}$ between consecutive propagation
layers. The propagation mechanism of a $K$-layer GCN is
\begin{equation}
\bar{\mathbf{X}}=\mathbf{H}^{(K)},\ \ \ \mathbf{H}^{(k)}=\mathrm{ReLU}\bigl(\mathbf{A}\mathbf{H}^{(k-1)}\mathbf{W}^{(k)}\bigr),\ \text{for}\ k=1,\ldots,K,\label{eq:GCN}
\end{equation}
where $\{\mathbf{W}^{(k)}\}_{k=1}^{K}$ are the weight matrices. The
relation between GCN and GSD is illustrated below.
\begin{prop}
\label{thm:GCN} A $K$-layer GCN model is equivalent to a $K$-layer
unrolled ProxGD network for solving a GSD problem \eqref{Formulation: GSD reg}
with $\alpha=0$ and $r(\mathbf{H})=\beta||\mathbf{H}||_{\mathbf{I}-\mathbf{T}_{\beta}}^{2}+I_{\{h_{ij}\geq0,\ \forall i,j\}}(\mathbf{H})$.
\end{prop}

Based on Proposition \ref{thm:GCN}, we can conclude that the GCN
model, owing to regularizer $r(\mathbf{H})$, does not merely perform
the target of global smoothing, which is also supported by some recent
works suggesting that oversmoothing does not necessarily happen in
deep GCN models \cite{yang2020revisiting,zhou2021understanding,cong2021provable}. 

\textbf{\textcolor[RGB]{128,42,42}{GCNII.}} The GCN model could suffer
from a severe performance degradation as it becomes deeper. To resolve
that performance degeneration issue, the GCNII model \cite{chen2020simple}
was proposed with two simple yet effective techniques, namely, initial
residual and identity mapping. A $K$-layer GCNII performs
\[
\bar{\mathbf{X}}=\mathbf{H}^{(K)},\ \ \ \mathbf{H}^{(k)}=\mathrm{ReLU}\Bigl(\bigl(\zeta\mathbf{A}\mathbf{H}^{(k-1)}+(1-\zeta)\mathbf{X}\bigr)\bigl(\xi\mathbf{W}^{(k)}+(1-\xi)\mathbf{I}\bigr)\Bigr),\ \text{for}\ k=1,\ldots,K,
\]
where $\zeta$ and $\xi$ are two prespecified parameters and $\mathbf{W}^{(k)}$
is a learnable matrix in the $k$-th layer. The relation between the
GCNII model and GSD problem is demonstrated in the following proposition.
\begin{prop}
\label{thm:GCNII} By concatenating $K$ unrolled ProxGD layers for
a GSD problem \eqref{Formulation: GSD reg} with $\beta=1-\alpha$,
$\mathbf{T}_{\alpha}=\mathbf{T}_{\beta}$, and $r(\mathbf{H})=\left\Vert \mathbf{H}\right\Vert _{\mathbf{I}-\mathbf{T}_{\beta}}^{2}\negthinspace+I_{\{h_{ij}\geq0,\ \forall i,j\}}(\mathbf{H})$,
we obtain a $K$-layer GCNII model. 
\end{prop}

Based on the results in Proposition \ref{thm:GCN} and Proposition
\ref{thm:GCNII} , we find that the underlying GSD objective for GCNII
has an additional signal fidelity term compared with that for GCN.
This could be used to support the empirical evidence from many studies
that the GCNII model is able to effectively relieve the performance
degradation problem encountered in deep GCNs \cite{chen2020simple,chen2021bag}.

\textbf{\textcolor[RGB]{128,42,42}{AirGNN.}} Besides inducing nonlinearity
to GNNs through imposing certain constraints $\mathcal{H}$ like the
non-negativity constraint, nonlinear FT steps can also be derived
through other ways. To make an example, we introduce the AirGNN model
\cite{liu2021graph}, which applies adaptive residual connections
for different nodes. A $K$-layer AirGNN is given as follows:
\begin{align*}
\bar{\mathbf{X}} & =\mathbf{H}^{(K)},\ \ \mathbf{h}_{i}^{(k)}=\mathrm{ReLU}\Bigl(1-\frac{1-\gamma}{2\gamma||[\mathbf{A}\mathbf{H}^{(k-1)}]_{i}-\mathbf{x}_{i}||_{2}}\Bigr)\bigl([\mathbf{A}\mathbf{H}^{(k-1)}]_{i}-\mathbf{x}_{i}\bigr)+\mathbf{x}_{i},\\
 & \hspace{10cm}\ \text{for}\ i\in\mathcal{V},\ k=1,\ldots,K,
\end{align*}
where $\mathbf{h}_{i}^{(k)}$ and $[\mathbf{A}\mathbf{H}^{(k-1)}]_{i}$
denote the $i$-th row of $\mathbf{H}^{(k)}$ and $\mathbf{A}\mathbf{H}^{(k-1)}$
respectively and $\gamma$ is a prespecified constant. The relation
between AirGNN and GSD is illustrated in the following proposition.
\begin{prop}
\label{thm:AirGNN} A $K$-layer AirGNN model is equivalent to a $K$-layer
unrolled ProxGD network for solving a GSD problem \eqref{Formulation: GSD reg}
with $\alpha=0$, $\mathbf{T}_{\alpha}=\mathbf{T}_{\beta}=\mathbf{I}$,
and $r(\mathbf{H})=(1-\beta)||\mathbf{H}-\mathbf{X}||_{2,1}$.
\end{prop}

The $||\mathbf{M}||_{2,1}$ denotes the $\ell_{2,1}$-norm of $\mathbf{M}$,
defined as $||\mathbf{M}||_{2,1}=\sum_{i}||\mathbf{m}_{i}||_{2}$
with $\mathbf{m}_{i}$ the $i$-th row of $\mathbf{M}$. It is generally
acknowledged that the $\ell_{1}$-norm is a robust loss function (corresponding
to a Laplacian distribution prior) in comparison with the $\ell_{2}$-norm
\cite{huber2004robust}. Therefore, the robustness of AirGNN against
abnormal node features can be readily established by investigating
its corresponding GSD problem.

So far, we have established the exact equivalence between a class
of popular GNN models with the unrolled GD/ProxGD networks for GSD
problems. We summarize all these connections in Table \ref{tab:GSD-GNN}.

\section{Proposed Model \label{Section: UGDGNN}}

For a GSD problem, if there is no prior knowledge on the signal, like
to be non-negative, imposing constraints may restrict the design space
and affect the solution quality. Besides, some recent works have empirically
demonstrated that the nonlinearities are the main factor for performance
degradations in deep GNNs, which hinders GNNs from learning node representations
from large receptive fields \cite{liu2020towards,zhou2021understanding}.
Moreover, some studies have theoretically shown that the nonlinearity
in the propagation layers will hurt the generalization ability of
GNNs \cite{cong2021provable}. Therefore, in this section, based on
the GSD problem in \eqref{Formulation: GSD reg}, we propose to build
a GNN via the unrolled GD layers.

In existing GNNs interpreted as unrolled GD networks, e.g., SGC, APPNP,
JKNet, and GPRGNN, our analyses have shown that the choices of the
model parameters $\eta^{(k)}$, $\alpha^{(k)}$, $\beta^{(k)}$, $\mathbf{T}_{\alpha}^{(k)}$,
and $\mathbf{T}_{\beta}^{(k)}$, are mostly hand-crafted and heuristic.
We conjecture that relieving such restrictions may help improving
the model performance. However, directly concatenating the unrolled
GD layers \eqref{Unrolling Step: GD} involves too many cumbersome
parameters which largely increases the model complexity. In the following,
we propose a novel GNN model unrolled from GD to mitigate this problem,
which is called UGDGNN. 
\begin{prop}
\label{UGDGNN} Concatenating $K$ unrolled GD layers \eqref{Unrolling Step: GD},
we obtain the following UGDGNN model:
\[
\bar{\mathbf{X}}=\sum_{k=0}^{K}\gamma^{(k)}\mathbf{A}^{k}\mathbf{X}\bigl(\zeta^{(k)}\mathbf{I}+\xi^{(k)}\mathbf{W}^{(k)}\bigr).
\]
\end{prop}

For a $K$-layer GNN, the nodes are able to aggregate information
from their $K$-th order neighborhood, which is to perform filtering
in the vertex domain \cite{shuman2013emerging}. Since the UGDGNN
model is derived from concatenating the general unrolled GD layers
\eqref{Unrolling Step: GD} without hand-crafted parameters, many
existing GNN models (e.g., SGC, APPNP, JKNet, and GPRGNN) can be regarded
as special cases of it, indicating the outstanding expressive power
of UGDGNN in vertex domain (detailed discussions are given in Appendix).
We further examine the expressive power of UGDGNN in the spectral
domain.
\begin{thm}
\label{thm: Spectral} A polynomial frequency filter of order $K$
is defined as $F_{K}(\mathbf{X})=\bigl(\sum_{k=0}^{K}\theta_{k}\mathbf{L}^{k}\bigr)\mathbf{X}$
with filtering coefficients $\theta_{k}$ for $k=0,\ldots,K$. A $K$-layer
UGDGNN model can express any \textup{$F_{K}(\mathbf{X})$,} i.e.,
a polynomial frequency filter of order $K$ with arbitrary coefficients.
\end{thm}

Based on Theorem \ref{thm: Spectral}, UGDGNN is able to express a
general polynomial graph filter and, hence, is capable of dealing
with different graph signal patterns (e.g., graph signals with either
high or low frequency components). By contrast, some existing GNNs
mentioned in the previous sections only have limited spectral expressiveness,
e.g., acting as a polynomial frequency filter with fixed coefficients
(see details in Appendix). Therefore, UGDGNN exhibits favorable expressive
powers in both the vertex domain and the spectral domain.

\section{Experiments}

\textbf{Datasets.} We conduct experiments on 7 datasets for semi-supervised
node classification tasks that are widely used to evaluate GNNs in
the literature, including three citation network datasets, i.e., Cora,
Citeseer, and Pubmed \cite{sen2008collective}, two co-authorship
datasets, i.e., Coauthor CS and Coauthor Physics \cite{shchur2018pitfalls},
a co-purchase dataset, i.e., Amazon Photo \cite{shchur2018pitfalls},
and a citation network collected in open graph benchmark (OGB), i.e.,
the OGBN-ArXiv dataset \cite{hu2020open}.

\textbf{Baselines.} We compare the proposed UGDGNN with representative
GNNs including GCN \cite{kipf2017semi}, SGC \cite{wu2019simplifying},
APPNP \cite{klicpera2019predict}, JKNet \cite{xu2018representation},
GCNII \cite{chen2020simple}, deep adaptive GNN (DAGNN) \cite{liu2020towards},
GPRGNN \cite{chien2021adaptive}.

\textbf{Node classification results.} The classification performance
of different GNN models on all benchmark datasets are summarized in
Table \ref{tab:Accuracy results}. Based on the results, we have the
following observations. 

Due to space limit, all the implementation details and hyperparameter
settings are given in Appendix.
\begin{itemize}
\item UGDGNN achieves superior or competitive performance over the state-of-the-art
GNNs on all the benchmark datasets. Such persuasive performance indicates
the promising prospects of designing new GNNs from the algorithm unrolling
perspective with the underlying GSDs.
\item GPRGNN consistently outperforms APPNP in all benchmark datasets, which
verifies the discovery that GPRGNN is a generalization of APPNP. Such
empirical evidence supports the rationality of interpreting GNNs from
the proposed unified algorithm unrolling perspective.
\item GNNs that act as frequency filters with fixed coefficients (i.e.,
GCN, SGC, APPNP) have overall inferior performance compared with GNNs
that have better expressive power (i.e., JKNet, GCNII, GPRGNN, UGDGNN),
validating efficacy of the expressive power measure.
\end{itemize}
\begin{table*}[t]
\begin{centering}
\caption{Classification performance (\%) on seven benchmark datasets. (\textbf{Bold}:
the best model; \uwave{wavy}: the runner-up model)\label{tab:Accuracy results}}
\par\end{centering}
\centering{}\resizebox{1 \textwidth}{!}{%
\begin{tabular}{c|ccccccc}
\hline 
\textbf{Model} & \textbf{Cora} & \textbf{CiteSeer} & \textbf{PubMed} & \textbf{CS} & \textbf{Physics} & \textbf{Photo} & \textbf{ArXiv}\tabularnewline
\hline 
GCN & \textcolor{black}{83.24}\textcolor{black}{\small{} $\pm$ 0.50} & 71.41{\small{} $\pm$ 0.63} & \textcolor{black}{79.10}\textcolor{black}{\small{} $\pm$ 0.42} & 91.76{\small{} $\pm$ 0.37} & 93.71{\small{} $\pm$ 0.67} & 90.37{\small{} $\pm$ 1.11} & 69.93{\small{} $\pm$ 0.11}\tabularnewline
SGC & \textcolor{black}{81.49}\textcolor{black}{\small{} $\pm$ 0.35} & \textcolor{black}{71.50}\textcolor{black}{\small{} $\pm$ 0.25} & \textcolor{black}{78.77}\textcolor{black}{\small{} $\pm$ 0.41} & \textcolor{black}{91.93}\textcolor{black}{\small{} $\pm$ 0.38} & \textcolor{black}{93.67}\textcolor{black}{\small{} $\pm$ 0.74} & \textcolor{black}{89.16}\textcolor{black}{\small{} $\pm$ 1.06} & 65.35{\small{} $\pm$ 0.10}\tabularnewline
APPNP & \textcolor{black}{82.88}\textcolor{black}{\small{} $\pm$ 0.25} & \textcolor{black}{71.53}\textcolor{black}{\small{} $\pm$ 0.59} & 79.10{\small{} $\pm$ 0.29} & \textcolor{black}{91.95}\textcolor{black}{\small{} $\pm$ 0.43} & \textcolor{black}{93.76}\textcolor{black}{\small{} $\pm$ 0.57} & 90.42{\small{} $\pm$ 0.64} & 69.84{\small{} $\pm$ 0.43}\tabularnewline
JKNet & \textcolor{black}{80.98}\textcolor{black}{\small{} $\pm$ 1.07} & 68.46{\small{} $\pm$ 1.37} & 76.35{\small{} $\pm$ 0.79} & \textcolor{black}{90.98}\textcolor{black}{\small{} $\pm$ 0.72} & 93.25{\small{} $\pm$ 0.82} & 87.55{\small{} $\pm$ 2.21} & 70.40{\small{} $\pm$ 0.19}\tabularnewline
GCNII & \textcolor{black}{\uwave{84.07}}\textcolor{black}{\small{}\uwave{
\mbox{$\pm$} 0.60}} & \textbf{72.32}\textbf{\small{} $\pm$ 0.61} & \textbf{79.56}\textbf{\small{} $\pm$ 0.42} & 91.03{\small{} $\pm$ 0.43} & 93.79{\small{} $\pm$ 0.65} & 80.01{\small{} $\pm$ 2.43} & 70.73{\small{} $\pm$ 0.43}\tabularnewline
DAGNN & \textcolor{black}{83.97}\textcolor{black}{\small{} $\pm$ 0.55} & 71.56{\small{} $\pm$ 0.48} & 79.11{\small{} $\pm$ 0.71} & \textcolor{black}{91.96}\textcolor{black}{\small{} $\pm$ 0.40} & \uwave{93.81}{\small{}\uwave{ \mbox{$\pm$} 0.84}} & 90.42{\small{} $\pm$ 1.26} & 70.67{\small{} $\pm$ 0.31}\tabularnewline
GPRGNN & \textcolor{black}{83.76}\textcolor{black}{\small{} $\pm$ 0.60} & \textcolor{black}{71.64}\textcolor{black}{\small{} $\pm$ 0.47} & 79.14{\small{} $\pm$ 0.63} & \uwave{92.03}{\small{}\uwave{ \mbox{$\pm$} 0.61}} & \textcolor{black}{93.75}\textcolor{black}{\small{} $\pm$ 0.72} & \textbf{91.63}\textbf{\small{} $\pm$ 0.95} & \uwave{70.82}{\small{}\uwave{ \mbox{$\pm$} 0.17}}\tabularnewline
\hline 
UGDGNN & \textbf{\textcolor{black}{84.15}}\textbf{\textcolor{black}{\small{}
$\pm$ 0.55}} & \uwave{71.68}{\small{}\uwave{ \mbox{$\pm$} 0.70}} & \uwave{79.22}{\small{}\uwave{ \mbox{$\pm$} 0.35}} & \textbf{92.18}\textbf{\small{} $\pm$ 0.66} & \textbf{94.14}\textbf{\small{} $\pm$ 0.50} & \uwave{90.49}{\small{}\uwave{ \mbox{$\pm$} 0.85}} & \textbf{70.91}\textbf{\small{} $\pm$ 0.22}\tabularnewline
\hline 
\end{tabular}}
\end{table*}

\begin{figure}[h]
\begin{centering}
\includegraphics[width=0.7\columnwidth]{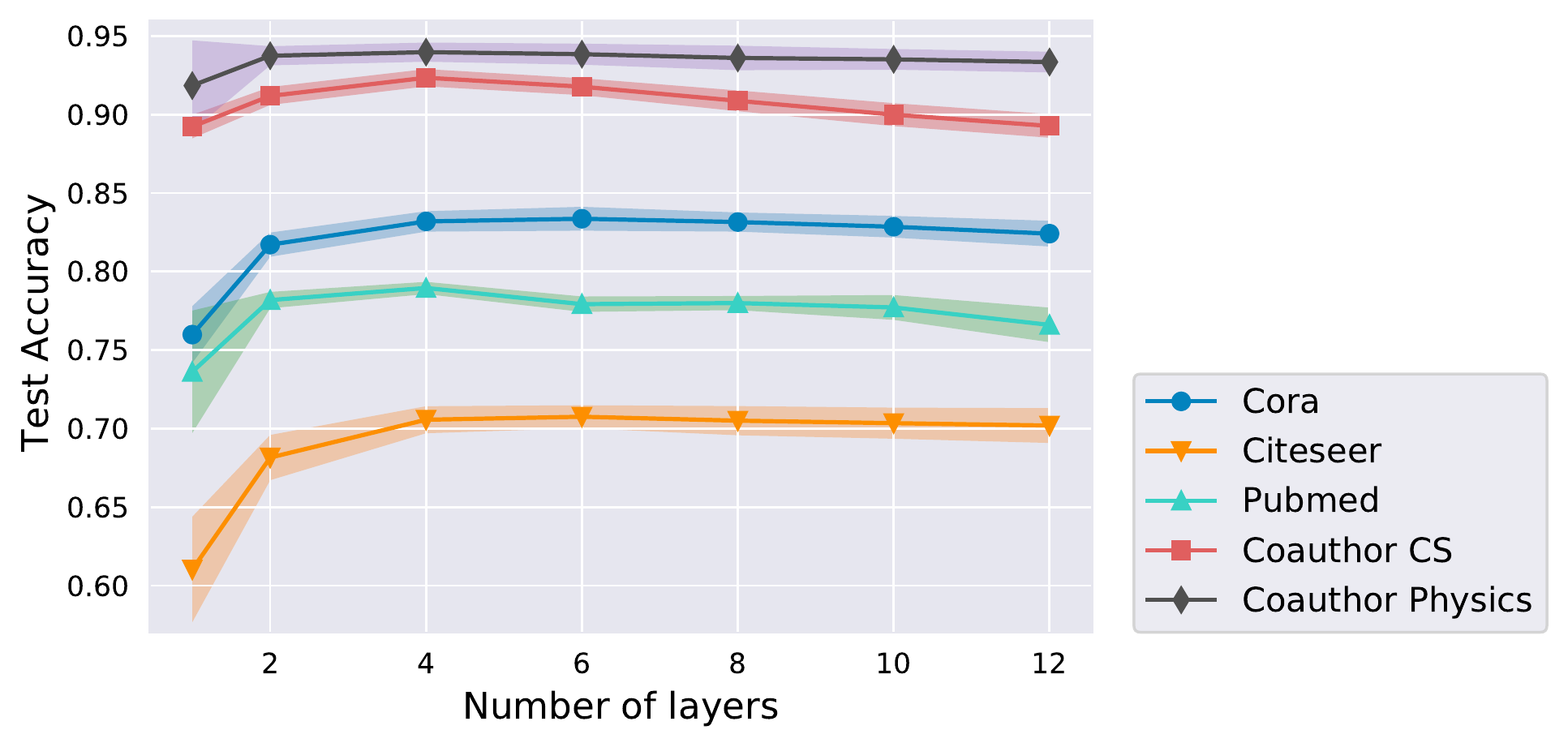}
\par\end{centering}
\caption{Classification performance of UGDGNN with different propagation depth.
\label{fig:fig}}
\end{figure}

\textbf{Propagation depth analysis.} We further investigate the impact
of propagation depth in UGDGNN, which is showcased in Figure \ref{fig:fig}.
From the figure, we can see that increasing the propagation depth
in UGDGNN can help to improve the classification performance. Moreover,
a UGDGNN with a small number of layers suffices to achieve a good
performance ($4\sim10$ layers on the benchmark datasets).

\section{Related Works\label{sec:Related-Work}}

With the development of various GNN models \cite{kipf2017semi,hamilton2017inductive,velivckovic2018graph,xu2018representation,wu2019simplifying,chen2020simple,chien2021adaptive},
understanding their working mechanisms has become an important research
filed. The graph signal processing tools \cite{dong2020graph}, which
have played important roles in some early designs of GNN architectures
\cite{bruna2014spectral,defferrard2016convolutional}, have been recently
used to analyze the properties \cite{ruiz2021graph} and to explain
the mechanisms \cite{fu2020understanding} of GNNs. 

There are some existing works intending to connect existing GNN models
with GSD optimization problems \cite{pan2020unified,zhu2021interpreting,ma2021unified}.
However, all of them focus on investigating the relationship among
different FA schemes of GNNs and neglect the indispensible FT steps
and none of them relates the GNNs to a bilevel optimization problem.
Therefore, their analyses can only be used to interpret one-layer
models or parameter-free models. By contrast, our work presents a
unified algorithm unrolling perspective on understanding the propagation
schemes (including both the FA and FT operations) of GNN models, which
is beyond the scope of the existing literatures and also brings much
more applications. 

Due to space limit, more discussions on related works and future works
are presented in Appendix.

\section{Conclusion}

In this paper, we have established the equivalence between a class
of GNN models and unrolled networks for solving GSD problems from
a unified algorithm unrolling perspective, with which deeper understandings
of the GNNs have been provided. We have shown that training these
GNNs is essentially to solve a bilevel optimization problem. Moreover,
a novel GNN model called UGDGNN has been proposed and is proved to
have favorable expressive power in both the vertex and the spectral
domain. Empirical results demonstrate that UGDGNN is able to achieve
superior or competitive performance over the state-of-the-art GNNs
on various benchmark datasets. The results indicate the promising
prospects of designing new models from the algorithm unrolling perspective
and leave much space for future explorations. 

\bibliographystyle{plainnat}
\bibliography{UGDGNN.bbl}

\begin{thebibliography}{74}
\providecommand{\natexlab}[1]{#1}
\providecommand{\url}[1]{\texttt{#1}}
\expandafter\ifx\csname urlstyle\endcsname\relax
  \providecommand{\doi}[1]{doi: #1}\else
  \providecommand{\doi}{doi: \begingroup \urlstyle{rm}\Url}\fi

\bibitem[Borgerding and Schniter(2016)]{borgerding2016onsager}
Mark Borgerding and Philip Schniter.
\newblock Onsager-corrected deep learning for sparse linear inverse problems.
\newblock In \emph{2016 IEEE Global Conference on Signal and Information
  Processing (GlobalSIP)}, pages 227--231. IEEE, 2016.

\bibitem[Bruna et~al.(2014)Bruna, Zaremba, Szlam, and LeCun]{bruna2014spectral}
Joan Bruna, Wojciech Zaremba, Arthur Szlam, and Yann LeCun.
\newblock Spectral networks and deep locally connected networks on graphs.
\newblock In \emph{International Conference on Learning Representations}, 2014.

\bibitem[Cai and Wang(2020)]{cai2020note}
Chen Cai and Yusu Wang.
\newblock A note on over-smoothing for graph neural networks.
\newblock \emph{arXiv preprint arXiv:2006.13318}, 2020.

\bibitem[Cappart et~al.(2021)Cappart, Ch{\'e}telat, Khalil, Lodi, Morris, and
  Veli{\v{c}}kovi{\'c}]{cappart2021combinatorial}
Quentin Cappart, Didier Ch{\'e}telat, Elias Khalil, Andrea Lodi, Christopher
  Morris, and Petar Veli{\v{c}}kovi{\'c}.
\newblock Combinatorial optimization and reasoning with graph neural networks.
\newblock \emph{arXiv preprint arXiv:2102.09544}, 2021.

\bibitem[Chen et~al.(2020)Chen, Wei, Huang, Ding, and Li]{chen2020simple}
Ming Chen, Zhewei Wei, Zengfeng Huang, Bolin Ding, and Yaliang Li.
\newblock Simple and deep graph convolutional networks.
\newblock In \emph{International Conference on Machine Learning}, pages
  1725--1735. PMLR, 2020.

\bibitem[Chen et~al.(2021{\natexlab{a}})Chen, Chen, Chen, Heaton, Liu, Wang,
  and Yin]{chen2021learning}
Tianlong Chen, Xiaohan Chen, Wuyang Chen, Howard Heaton, Jialin Liu, Zhangyang
  Wang, and Wotao Yin.
\newblock Learning to optimize: A primer and a benchmark.
\newblock \emph{arXiv preprint arXiv:2103.12828}, 2021{\natexlab{a}}.

\bibitem[Chen et~al.(2021{\natexlab{b}})Chen, Zhou, Duan, Zheng, Wang, Hu, and
  Wang]{chen2021bag}
Tianlong Chen, Kaixiong Zhou, Keyu Duan, Wenqing Zheng, Peihao Wang, Xia Hu,
  and Zhangyang Wang.
\newblock Bag of tricks for training deeper graph neural networks: A
  comprehensive benchmark study.
\newblock \emph{arXiv preprint arXiv:2108.10521}, 2021{\natexlab{b}}.

\bibitem[Chen et~al.(2018)Chen, Liu, Wang, and Yin]{chen2018theoretical}
Xiaohan Chen, Jialin Liu, Zhangyang Wang, and Wotao Yin.
\newblock Theoretical linear convergence of unfolded {ISTA} and its practical
  weights and thresholds.
\newblock In \emph{Advances in Neural Information Processing Systems}, pages
  9079--9089, 2018.

\bibitem[Chien et~al.(2021)Chien, Peng, Li, and Milenkovic]{chien2021adaptive}
Eli Chien, Jianhao Peng, Pan Li, and Olgica Milenkovic.
\newblock Adaptive universal generalized pagerank graph neural network.
\newblock In \emph{International Conference on Learning Representations}, 2021.

\bibitem[Combettes and Pesquet(2020)]{combettes2020deep}
Patrick~L Combettes and Jean-Christophe Pesquet.
\newblock Deep neural network structures solving variational inequalities.
\newblock \emph{Set-Valued and Variational Analysis}, pages 1--28, 2020.

\bibitem[Cong et~al.(2021)Cong, Ramezani, and Mahdavi]{cong2021provable}
Weilin Cong, Morteza Ramezani, and Mehrdad Mahdavi.
\newblock On provable benefits of depth in training graph convolutional
  networks.
\newblock \emph{Advances in Neural Information Processing Systems}, 34, 2021.

\bibitem[Defferrard et~al.(2016)Defferrard, Bresson, and
  Vandergheynst]{defferrard2016convolutional}
Micha{\"e}l Defferrard, Xavier Bresson, and Pierre Vandergheynst.
\newblock Convolutional neural networks on graphs with fast localized spectral
  filtering.
\newblock \emph{Advances in Neural Information Processing Systems},
  29:\penalty0 3844--3852, 2016.

\bibitem[Dempe and Zemkoho(2020)]{dempe2020bilevel}
Stephan Dempe and Alain Zemkoho.
\newblock \emph{Bilevel optimization}.
\newblock Springer, 2020.

\bibitem[Dong et~al.(2020)Dong, Thanou, Toni, Bronstein, and
  Frossard]{dong2020graph}
Xiaowen Dong, Dorina Thanou, Laura Toni, Michael Bronstein, and Pascal
  Frossard.
\newblock Graph signal processing for machine learning: A review and new
  perspectives.
\newblock \emph{IEEE Signal Processing Magazine}, 37\penalty0 (6):\penalty0
  117--127, 2020.

\bibitem[Fan et~al.(2021)Fan, Liu, Jin, Zhao, Tang, and Li]{fan2021graph}
Wenqi Fan, Xiaorui Liu, Wei Jin, Xiangyu Zhao, Jiliang Tang, and Qing Li.
\newblock Graph trend networks for recommendations.
\newblock \emph{arXiv preprint arXiv:2108.05552}, 2021.

\bibitem[Fey and Lenssen(2019)]{Fey/Lenssen/2019}
Matthias Fey and Jan~E. Lenssen.
\newblock Fast graph representation learning with {PyTorch Geometric}.
\newblock In \emph{ICLR Workshop on Representation Learning on Graphs and
  Manifolds}, 2019.

\bibitem[Fu et~al.(2020)Fu, Hou, Zhang, Ma, Kamhoua, and
  Cheng]{fu2020understanding}
Guoji Fu, Yifan Hou, Jian Zhang, Kaili Ma, Barakeel~Fanseu Kamhoua, and James
  Cheng.
\newblock Understanding graph neural networks from graph signal denoising
  perspectives.
\newblock \emph{arXiv preprint arXiv:2006.04386}, 2020.

\bibitem[Fu et~al.(2021)Fu, Zhao, and Bian]{fu2021p}
Guoji Fu, Peilin Zhao, and Yatao Bian.
\newblock $ p $-laplacian based graph neural networks.
\newblock \emph{arXiv preprint arXiv:2111.07337}, 2021.

\bibitem[Gilmer et~al.(2017)Gilmer, Schoenholz, Riley, Vinyals, and
  Dahl]{gilmer2017neural}
Justin Gilmer, Samuel~S Schoenholz, Patrick~F Riley, Oriol Vinyals, and
  George~E Dahl.
\newblock Neural message passing for quantum chemistry.
\newblock In \emph{International Conference on Machine Learning}, pages
  1263--1272. PMLR, 2017.

\bibitem[Gregor and LeCun(2010)]{gregor2010learning}
Karol Gregor and Yann LeCun.
\newblock Learning fast approximations of sparse coding.
\newblock In \emph{International Conference on Machine Learning}, pages
  399--406, 2010.

\bibitem[Hamilton et~al.(2017)Hamilton, Ying, and
  Leskovec]{hamilton2017inductive}
William~L Hamilton, Rex Ying, and Jure Leskovec.
\newblock Inductive representation learning on large graphs.
\newblock In \emph{Advances in Neural Information Processing Systems}, pages
  1025--1035, 2017.

\bibitem[Hershey et~al.(2014)Hershey, Roux, and Weninger]{hershey2014deep}
John~R Hershey, Jonathan~Le Roux, and Felix Weninger.
\newblock Deep unfolding: Model-based inspiration of novel deep architectures.
\newblock \emph{arXiv preprint arXiv:1409.2574}, 2014.

\bibitem[Hoang et~al.(2021)Hoang, Maehara, and Murata]{hoang2021revisiting}
NT~Hoang, Takanori Maehara, and Tsuyoshi Murata.
\newblock Revisiting graph neural networks: Graph filtering perspective.
\newblock In \emph{International Conference on Pattern Recognition}, pages
  8376--8383. IEEE, 2021.

\bibitem[Hu et~al.(2020)Hu, Fey, Zitnik, Dong, Ren, Liu, Catasta, and
  Leskovec]{hu2020open}
Weihua Hu, Matthias Fey, Marinka Zitnik, Yuxiao Dong, Hongyu Ren, Bowen Liu,
  Michele Catasta, and Jure Leskovec.
\newblock Open graph benchmark: Datasets for machine learning on graphs.
\newblock \emph{Advances in Neural Information Processing Systems}, 2020.

\bibitem[Huber(2004)]{huber2004robust}
Peter~J Huber.
\newblock \emph{Robust statistics}, volume 523.
\newblock John Wiley \& Sons, 2004.

\bibitem[Jiang et~al.(2022)Jiang, Han, Fan, Liu, Zou, Mostafavi, and
  Hu]{jiang2022fmp}
Zhimeng Jiang, Xiaotian Han, Chao Fan, Zirui Liu, Na~Zou, Ali Mostafavi, and
  Xia Hu.
\newblock Fmp: Toward fair graph message passing against topology bias.
\newblock \emph{arXiv preprint arXiv:2202.04187}, 2022.

\bibitem[Kalofolias(2016)]{kalofolias2016learn}
Vassilis Kalofolias.
\newblock How to learn a graph from smooth signals.
\newblock In \emph{Artificial Intelligence and Statistics}, pages 920--929.
  PMLR, 2016.

\bibitem[Kingma and Ba(2015)]{kingma2015adam}
Diederik~P Kingma and Jimmy Ba.
\newblock Adam: A method for stochastic optimization.
\newblock In \emph{International Conference on Learning Representations
  (ICLR)}, 2015.

\bibitem[Kipf and Welling(2017)]{kipf2017semi}
Thomas~N. Kipf and Max Welling.
\newblock Semi-supervised classification with graph convolutional networks.
\newblock In \emph{International Conference on Learning Representations}, 2017.

\bibitem[Klicpera et~al.(2019)Klicpera, Bojchevski, and
  G{\"u}nnemann]{klicpera2019predict}
Johannes Klicpera, Aleksandar Bojchevski, and Stephan G{\"u}nnemann.
\newblock Predict then propagate: Graph neural networks meet personalized
  pagerank.
\newblock In \emph{International Conference on Learning Representations}, 2019.

\bibitem[Kloumann et~al.(2017)Kloumann, Ugander, and
  Kleinberg]{kloumann2017block}
Isabel~M Kloumann, Johan Ugander, and Jon Kleinberg.
\newblock Block models and personalized pagerank.
\newblock \emph{Proceedings of the National Academy of Sciences}, 114\penalty0
  (1):\penalty0 33--38, 2017.

\bibitem[Li et~al.(2019)Li, Chien, and Milenkovic]{li2019optimizing}
Pan Li, I~Chien, and Olgica Milenkovic.
\newblock Optimizing generalized pagerank methods for seed-expansion community
  detection.
\newblock \emph{Advances in Neural Information Processing Systems},
  32:\penalty0 11710--11721, 2019.

\bibitem[Li et~al.(2018)Li, Han, and Wu]{li2018deeper}
Qimai Li, Zhichao Han, and Xiao-Ming Wu.
\newblock Deeper insights into graph convolutional networks for semi-supervised
  learning.
\newblock In \emph{AAAI Conference on Artificial Intelligence}, 2018.

\bibitem[Liu and Chen(2019)]{liu2019alista}
Jialin Liu and Xiaohan Chen.
\newblock Alista: Analytic weights are as good as learned weights in lista.
\newblock In \emph{International Conference on Learning Representations
  (ICLR)}, 2019.

\bibitem[Liu et~al.(2020)Liu, Gao, and Ji]{liu2020towards}
Meng Liu, Hongyang Gao, and Shuiwang Ji.
\newblock Towards deeper graph neural networks.
\newblock In \emph{ACM SIGKDD International Conference on Knowledge Discovery
  \& Data Mining}, pages 338--348, 2020.

\bibitem[Liu et~al.(2021{\natexlab{a}})Liu, Ding, Jin, Xu, Ma, Liu, and
  Tang]{liu2021graph}
Xiaorui Liu, Jiayuan Ding, Wei Jin, Han Xu, Yao Ma, Zitao Liu, and Jiliang
  Tang.
\newblock Graph neural networks with adaptive residual.
\newblock In \emph{Advances in Neural Information Processing Systems},
  2021{\natexlab{a}}.

\bibitem[Liu et~al.(2021{\natexlab{b}})Liu, Jin, Ma, Li, Liu, Wang, Yan, and
  Tang]{liu2021elastic}
Xiaorui Liu, Wei Jin, Yao Ma, Yaxin Li, Hua Liu, Yiqi Wang, Ming Yan, and
  Jiliang Tang.
\newblock Elastic graph neural networks.
\newblock In \emph{International Conference on Machine Learning}, pages
  6837--6849. PMLR, 2021{\natexlab{b}}.

\bibitem[Ma et~al.(2021)Ma, Liu, Zhao, Liu, Tang, and Shah]{ma2021unified}
Yao Ma, Xiaorui Liu, Tong Zhao, Yozen Liu, Jiliang Tang, and Neil Shah.
\newblock A unified view on graph neural networks as graph signal denoising.
\newblock In \emph{ACM International Conference on Information \& Knowledge
  Management}, pages 1202--1211, 2021.

\bibitem[Monga et~al.(2021)Monga, Li, and Eldar]{monga2021algorithm}
Vishal Monga, Yuelong Li, and Yonina~C Eldar.
\newblock Algorithm unrolling: Interpretable, efficient deep learning for
  signal and image processing.
\newblock \emph{IEEE Signal Processing Magazine}, 38\penalty0 (2):\penalty0
  18--44, 2021.

\bibitem[Newman(2018)]{newman2018networks}
Mark Newman.
\newblock \emph{Networks}.
\newblock Oxford University Press, 2018.

\bibitem[Page et~al.(1998)Page, Brin, Motwani, and Winograd]{page1998pagerank}
Lawrence Page, Sergey Brin, Rajeev Motwani, and Terry Winograd.
\newblock The pagerank citation ranking: Bringing order to the web.
\newblock Technical report, Stanford InfoLab, 1998.

\bibitem[Pan et~al.(2020)Pan, Song, and Huang]{pan2020unified}
Xuran Pan, Shiji Song, and Gao Huang.
\newblock A unified framework for convolution-based graph neural networks.
\newblock \emph{https://openreview.net/forum?id=zUMD--Fb9Bt}, 2020.

\bibitem[Papyan et~al.(2017)Papyan, Romano, and Elad]{papyan2017convolutional}
Vardan Papyan, Yaniv Romano, and Michael Elad.
\newblock Convolutional neural networks analyzed via convolutional sparse
  coding.
\newblock \emph{The Journal of Machine Learning Research}, 18\penalty0
  (1):\penalty0 2887--2938, 2017.

\bibitem[Pei et~al.(2020)Pei, Wei, Chang, Lei, and Yang]{pei2020geom}
Hongbin Pei, Bingzhe Wei, Kevin Chen-Chuan Chang, Yu~Lei, and Bo~Yang.
\newblock Geom-{GCN}: Geometric graph convolutional networks.
\newblock In \emph{International Conference on Learning Representations}, 2020.

\bibitem[Rong et~al.(2020)Rong, Bian, Xu, Xie, Wei, Huang, and
  Huang]{rong2020self}
Yu~Rong, Yatao Bian, Tingyang Xu, Weiyang Xie, Ying Wei, Wenbing Huang, and
  Junzhou Huang.
\newblock Self-supervised graph transformer on large-scale molecular data.
\newblock \emph{Advances in Neural Information Processing Systems},
  33:\penalty0 12559--12571, 2020.

\bibitem[Ruiz et~al.(2021)Ruiz, Gama, and Ribeiro]{ruiz2021graph}
Luana Ruiz, Fernando Gama, and Alejandro Ribeiro.
\newblock Graph neural networks: Architectures, stability, and transferability.
\newblock \emph{Proceedings of the IEEE}, 109\penalty0 (5):\penalty0 660--682,
  2021.

\bibitem[Satorras and Estrach(2018)]{satorras2018few}
Victor~Garcia Satorras and Joan~Bruna Estrach.
\newblock Few-shot learning with graph neural networks.
\newblock In \emph{International Conference on Learning Representations}, 2018.

\bibitem[Sen et~al.(2008)Sen, Namata, Bilgic, Getoor, Galligher, and
  Eliassi-Rad]{sen2008collective}
Prithviraj Sen, Galileo Namata, Mustafa Bilgic, Lise Getoor, Brian Galligher,
  and Tina Eliassi-Rad.
\newblock Collective classification in network data.
\newblock \emph{AI Magazine}, 29\penalty0 (3):\penalty0 93--93, 2008.

\bibitem[Shchur et~al.(2018)Shchur, Mumme, Bojchevski, and
  G{\"u}nnemann]{shchur2018pitfalls}
Oleksandr Shchur, Maximilian Mumme, Aleksandar Bojchevski, and Stephan
  G{\"u}nnemann.
\newblock Pitfalls of graph neural network evaluation.
\newblock \emph{arXiv preprint arXiv:1811.05868}, 2018.

\bibitem[Shuman et~al.(2013)Shuman, Narang, Frossard, Ortega, and
  Vandergheynst]{shuman2013emerging}
David~I Shuman, Sunil~K Narang, Pascal Frossard, Antonio Ortega, and Pierre
  Vandergheynst.
\newblock The emerging field of signal processing on graphs: Extending
  high-dimensional data analysis to networks and other irregular domains.
\newblock \emph{IEEE Signal Processing Magazine}, 30\penalty0 (3):\penalty0
  83--98, 2013.

\bibitem[Srebro and Jaakkola(2003)]{srebro2003weighted}
Nathan Srebro and Tommi Jaakkola.
\newblock Weighted low-rank approximations.
\newblock In \emph{International Conference on Machine Learning}, pages
  720--727, 2003.

\bibitem[Stankovi{\'c} et~al.(2019)Stankovi{\'c}, Dakovi{\'c}, and
  Sejdi{\'c}]{stankovic2019introduction}
Ljubi{\v{s}}a Stankovi{\'c}, Milo{\v{s}} Dakovi{\'c}, and Ervin Sejdi{\'c}.
\newblock Introduction to graph signal processing.
\newblock In \emph{Vertex-Frequency Analysis of Graph Signals}, pages 3--108.
  Springer, 2019.

\bibitem[Tolstaya et~al.(2020)Tolstaya, Gama, Paulos, Pappas, Kumar, and
  Ribeiro]{tolstaya2020learning}
Ekaterina Tolstaya, Fernando Gama, James Paulos, George Pappas, Vijay Kumar,
  and Alejandro Ribeiro.
\newblock Learning decentralized controllers for robot swarms with graph neural
  networks.
\newblock In \emph{Conference on Robot Learning}, pages 671--682. PMLR, 2020.

\bibitem[Veli{\v{c}}kovi{\'c} et~al.(2018)Veli{\v{c}}kovi{\'c}, Cucurull,
  Casanova, Romero, Li{\`o}, and Bengio]{velivckovic2018graph}
Petar Veli{\v{c}}kovi{\'c}, Guillem Cucurull, Arantxa Casanova, Adriana Romero,
  Pietro Li{\`o}, and Yoshua Bengio.
\newblock Graph attention networks.
\newblock In \emph{International Conference on Learning Representations}, 2018.

\bibitem[Wang et~al.(2015)Wang, Sharpnack, Smola, and
  Tibshirani]{wang2015trend}
Yu-Xiang Wang, James Sharpnack, Alex Smola, and Ryan Tibshirani.
\newblock Trend filtering on graphs.
\newblock In \emph{Artificial Intelligence and Statistics}, pages 1042--1050.
  PMLR, 2015.

\bibitem[Wu et~al.(2019)Wu, Souza, Zhang, Fifty, Yu, and
  Weinberger]{wu2019simplifying}
Felix Wu, Amauri Souza, Tianyi Zhang, Christopher Fifty, Tao Yu, and Kilian
  Weinberger.
\newblock Simplifying graph convolutional networks.
\newblock In \emph{International Conference on Machine Learning}, pages
  6861--6871. PMLR, 2019.

\bibitem[Wu et~al.(2021)Wu, Chen, Shen, Guo, Gao, Li, Pei, and
  Long]{wu2021graph}
Lingfei Wu, Yu~Chen, Kai Shen, Xiaojie Guo, Hanning Gao, Shucheng Li, Jian Pei,
  and Bo~Long.
\newblock Graph neural networks for natural language processing: A survey.
\newblock \emph{arXiv preprint arXiv:2106.06090}, 2021.

\bibitem[Wu et~al.(2022)Wu, Cui, Pei, and Zhao]{GNNBook2022}
Lingfei Wu, Peng Cui, Jian Pei, and Liang Zhao.
\newblock \emph{Graph Neural Networks: Foundations, Frontiers, and
  Applications}.
\newblock Springer, Singapore, 2022.

\bibitem[Wu et~al.(2020)Wu, Pan, Chen, Long, Zhang, and
  Philip]{wu2020comprehensive}
Zonghan Wu, Shirui Pan, Fengwen Chen, Guodong Long, Chengqi Zhang, and S~Yu
  Philip.
\newblock A comprehensive survey on graph neural networks.
\newblock \emph{IEEE Transactions on Neural Networks and Learning Systems},
  32\penalty0 (1):\penalty0 4--24, 2020.

\bibitem[Xu et~al.(2018)Xu, Li, Tian, Sonobe, Kawarabayashi, and
  Jegelka]{xu2018representation}
Keyulu Xu, Chengtao Li, Yonglong Tian, Tomohiro Sonobe, Ken-ichi Kawarabayashi,
  and Stefanie Jegelka.
\newblock Representation learning on graphs with jumping knowledge networks.
\newblock In \emph{International Conference on Machine Learning}, pages
  5453--5462. PMLR, 2018.

\bibitem[Yang et~al.(2020)Yang, Wang, Yao, Liu, and
  Abdelzaher]{yang2020revisiting}
Chaoqi Yang, Ruijie Wang, Shuochao Yao, Shengzhong Liu, and Tarek Abdelzaher.
\newblock Revisiting over-smoothing in deep {GCNs}.
\newblock \emph{arXiv preprint arXiv:2003.13663}, 2020.

\bibitem[Yang et~al.(2021{\natexlab{a}})Yang, Ma, and
  Cheng]{yang2021rethinking}
Han Yang, Kaili Ma, and James Cheng.
\newblock Rethinking graph regularization for graph neural networks.
\newblock In \emph{AAAI Conference on Artificial Intelligence}, volume~35,
  pages 4573--4581, 2021{\natexlab{a}}.

\bibitem[Yang et~al.(2021{\natexlab{b}})Yang, Liu, Wang, Zhou, Gan, Wei, Zhang,
  Huang, and Wipf]{yang2021graph}
Yongyi Yang, Tang Liu, Yangkun Wang, Jinjing Zhou, Quan Gan, Zhewei Wei, Zheng
  Zhang, Zengfeng Huang, and David Wipf.
\newblock Graph neural networks inspired by classical iterative algorithms.
\newblock \emph{arXiv preprint arXiv:2103.06064}, 2021{\natexlab{b}}.

\bibitem[Yang et~al.(2021{\natexlab{c}})Yang, Wang, Huang, and
  Wipf]{yang2021implicit}
Yongyi Yang, Yangkun Wang, Zengfeng Huang, and David Wipf.
\newblock Implicit vs unfolded graph neural networks.
\newblock \emph{arXiv preprint arXiv:2111.06592}, 2021{\natexlab{c}}.

\bibitem[Yang et~al.(2016)Yang, Cohen, and Salakhudinov]{yang2016revisiting}
Zhilin Yang, William Cohen, and Ruslan Salakhudinov.
\newblock Revisiting semi-supervised learning with graph embeddings.
\newblock In \emph{International Conference on Machine Learning}, pages 40--48.
  PMLR, 2016.

\bibitem[Ying et~al.(2018)Ying, He, Chen, Eksombatchai, Hamilton, and
  Leskovec]{ying2018graph}
Rex Ying, Ruining He, Kaifeng Chen, Pong Eksombatchai, William~L Hamilton, and
  Jure Leskovec.
\newblock Graph convolutional neural networks for web-scale recommender
  systems.
\newblock In \emph{Proceedings of the 24th ACM SIGKDD International Conference
  on Knowledge Discovery \& Data Mining}, pages 974--983, 2018.

\bibitem[You et~al.(2018)You, Liu, Ying, Pande, and Leskovec]{you2018graph}
Jiaxuan You, Bowen Liu, Rex Ying, Vijay Pande, and Jure Leskovec.
\newblock Graph convolutional policy network for goal-directed molecular graph
  generation.
\newblock In \emph{Proceedings of the 32nd International Conference on Neural
  Information Processing Systems}, pages 6412--6422, 2018.

\bibitem[You et~al.(2019)You, Ying, and Leskovec]{you2019position}
Jiaxuan You, Rex Ying, and Jure Leskovec.
\newblock Position-aware graph neural networks.
\newblock In \emph{International Conference on Machine Learning}, pages
  7134--7143. PMLR, 2019.

\bibitem[Zhang et~al.(2020{\natexlab{a}})Zhang, Yan, Xie, Xia, and
  Zhang]{zhang2020revisiting}
Hongwei Zhang, Tijin Yan, Zenjun Xie, Yuanqing Xia, and Yuan Zhang.
\newblock Revisiting graph convolutional network on semi-supervised node
  classification from an optimization perspective.
\newblock \emph{arXiv preprint arXiv:2009.11469}, 2020{\natexlab{a}}.

\bibitem[Zhang et~al.(2020{\natexlab{b}})Zhang, Cui, and Zhu]{zhang2020deep}
Ziwei Zhang, Peng Cui, and Wenwu Zhu.
\newblock Deep learning on graphs: A survey.
\newblock \emph{IEEE Transactions on Knowledge and Data Engineering},
  2020{\natexlab{b}}.

\bibitem[Zhao and Akoglu(2020)]{Model:PairNorm}
Lingxiao Zhao and Leman Akoglu.
\newblock {PairNorm}: Tackling oversmoothing in {GNN}s.
\newblock In \emph{International Conference on Learning Representations}, 2020.

\bibitem[Zhou et~al.(2020)Zhou, Cui, Hu, Zhang, Yang, Liu, Wang, Li, and
  Sun]{zhou2020graph}
Jie Zhou, Ganqu Cui, Shengding Hu, Zhengyan Zhang, Cheng Yang, Zhiyuan Liu,
  Lifeng Wang, Changcheng Li, and Maosong Sun.
\newblock Graph neural networks: A review of methods and applications.
\newblock \emph{AI Open}, 1:\penalty0 57--81, 2020.

\bibitem[Zhou et~al.(2021)Zhou, Dong, Wang, Lee, Hooi, Xu, and
  Feng]{zhou2021understanding}
Kuangqi Zhou, Yanfei Dong, Kaixin Wang, Wee~Sun Lee, Bryan Hooi, Huan Xu, and
  Jiashi Feng.
\newblock Understanding and resolving performance degradation in deep graph
  convolutional networks.
\newblock In \emph{ACM International Conference on Information \& Knowledge
  Management}, pages 2728--2737, 2021.

\bibitem[Zhu et~al.(2021)Zhu, Wang, Shi, Ji, and Cui]{zhu2021interpreting}
Meiqi Zhu, Xiao Wang, Chuan Shi, Houye Ji, and Peng Cui.
\newblock Interpreting and unifying graph neural networks with an optimization
  framework.
\newblock In \emph{Web Conference 2021}, pages 1215--1226, 2021.

\end{thebibliography}

\section*{Checklist}
\begin{enumerate}

\item For all authors...
\begin{enumerate}   
\item Do the main claims made in the abstract and introduction accurately reflect the paper's contributions and scope?     \answerYes{}   
\item Did you describe the limitations of your work?     \answerYes{}   \item Did you discuss any potential negative societal impacts of your work?     \answerYes{}  
\item Have you read the ethics review guidelines and ensured that your paper conforms to them?     \answerYes{}  
\end{enumerate}

\item If you are including theoretical results... 
\begin{enumerate}   
\item Did you state the full set of assumptions of all theoretical results?     \answerYes{}          
\item Did you include complete proofs of all theoretical results?     \answerYes{}  
\end{enumerate}

\item If you ran experiments... 
\begin{enumerate}   
\item Did you include the code, data, and instructions needed to reproduce the main experimental results (either in the supplemental material or as a URL)?     \answerYes{} 
\item Did you specify all the training details (e.g., data splits, hyperparameters, how they were chosen)?     \answerYes{}         
\item Did you report error bars (e.g., with respect to the random seed after running experiments multiple times)?     \answerYes{}          \item Did you include the total amount of compute and the type of resources used (e.g., type of GPUs, internal cluster, or cloud provider)?     \answerYes{}  
\end{enumerate}

\item If you are using existing assets (e.g., code, data, models) or curating/releasing new assets... 
\begin{enumerate}   
\item If your work uses existing assets, did you cite the creators?     \answerYes{}  
\item Did you mention the license of the assets?     \answerNo{}The assets are open to academic use.
\item Did you include any new assets either in the supplemental material or as a URL?     \answerNo{}
\item Did you discuss whether and how consent was obtained from people whose data you're using/curating?     \answerNA{}   
\item Did you discuss whether the data you are using/curating contains personally identifiable information or offensive content?     \answerNA{} \end{enumerate}

\item If you used crowdsourcing or conducted research with human subjects... 
\begin{enumerate}   
\item Did you include the full text of instructions given to participants and screenshots, if applicable?     \answerNA{} 
\item Did you describe any potential participant risks, with links to Institutional Review Board (IRB) approvals, if applicable?     \answerNA{}   
\item Did you include the estimated hourly wage paid to participants and the total amount spent on participant compensation?     \answerNA{} 
\end{enumerate}
\end{enumerate}\newpage{}

\appendix
\renewcommand \thepart{} 
\renewcommand \partname{}

\doparttoc 
\faketableofcontents 
\addcontentsline{toc}{section}{Appendix} 
\part{Appendix} 
\parttoc

\section{Additional Related Works and Future Directions}

Several earlier papers found that the FA step in GCN \cite{kipf2017semi}
and SGC \cite{wu2019simplifying} can be seen as the outcome of a
special signal smoothing technique called Laplacian smoothing \cite{li2018deeper,hoang2021revisiting,yang2021rethinking}.
Inspired by this observation, there is a line of research focusing
on designing novel GNN models based on a ``graph regularized optimization
problem,'' which actually is the vanilla form of the GSD problem given
in \eqref{Formulation: GSD}. If a iterative numerical algorithm (e.g.,
the gradient descent algorithm, matrix inversion algorithm, etc.)
is applied to solve this problem, stacking all the iterative steps
will naturally lead to a computational model. Since computation over
graphs with consecutive propagation layers are involved, such models
are all named GNN models. This idea of GNN design (i.e., designing
GNNs from iterative algorithms) looks similar to the one proposed
in this paper in that both of them originate from GSD optimization
problems. Actually they are different in that the GNN models designed
from the vanilla GSD problem will finally lead to GNN models without
any parameters (except possible parameters included in $p_{\mathsf{pre}}$
and $p_{\mathsf{pos}}$). In other words, no parameters need to be
trained in the feature propagation layers in these models. In this
paper, we considered a class of popular GNN models which are not originally
motivated from the vanilla GSD problem and have learnable parameters
in the feature propagation layers. While empirical results have propelled
GNN to new heights in recent years, the philosophy behind them is
elusive. We firmly believe that practice grounded in theory could
help accelerate the GNN research and possibly lead to the discovery
of new fields we cannot even conceive of yet. The original intention
of this paper is actually to launch an attempt in this direction.
We first realized the GSD problem in the area of signal processing
over graph shares the same spirit as the GNN model in the area of
machine learning over graph. Then, such a connection can actually
be made clear through the algorithm unrolling perspective. To the
best of our knowledge, this paper is the first attempt of the kind
in the literature towards obtaining a holistic understanding of the
working mechanisms of existing GNN models.

Among the literature on GNN models designed from the ``graph regularized
optimization problem,'' there are research studies targeting at modifying
the signal smoothing term in the optimization problem to strengthen
the capability of GNNs. For example, inspired by the idea of trend
filtering \cite{wang2015trend}, authors in \cite{liu2021elastic}
and \cite{fan2021graph} proposed to replace the Laplacian smoothing
term (in the form of $\ell_{2}$ norm) with an $\ell_{2,1}$ norm
to promote robustness against abnormal edges. Also for robustness
pursuit, authors in \cite{yang2021graph} replaced the Laplacian smoothing
term with some robustness promoting nonlinear functions over pairwise
node distances. Besides promoting smoothness over connected nodes,
\cite{zhang2020revisiting,Model:PairNorm} suggested to further promoting
the non-smoothness over the disconnected nodes, which is achieved
by deducting the sum of distances between disconnected pairs of nodes
from the GSD objective. Moreover, \cite{jiang2022fmp} augments the
GSD objective with a fairness term to fight against large topology
bias. Most recently, \cite{fu2021p} proposed $p$-Laplacian message
passing and $^{p}$GNN, which is capable of dealing with heterophilic
graphs and is robust to noisy edges. However, all these techniques
amount to designing the FA steps for GNNs, neglecting the power of
FT steps, which limits the representation ability of GNNs. 

With the proposed algorithm unrolling perspective for understanding
GNNs, a wider range of GNNs can be interpreted as solving properly
specified GSD problems via unrolled networks. Such an interpretation
can inspire more powerful GNN designs that may contain FT steps. Specially,
new GNN models can be derived based on other variants of GSD objectives
(possibly non-convex or with non-Euclidean structure), other iterative
algorithms for problem resolution (e.g.,  numerical optimization algorithms,
numerical linear algebra algorithms, etc.), other unrolling schemes,
and so on. Besides, more characteristics of the GNNs can be revealed
by further investigating the properties of the underlying GSDs, the
iterative algorithms, and/or the unrolling schemes. 

\section{Societal Impact}

The GNN model, a graph-based machine learning method, has achieved
great success in many application fields. The unified algorithm unrolling
perspective proposed in this paper will benefit the society as it
provides a fresh view for understanding these models and can motivate
new model designs. For example, one can choose to design a more robust
GNN model via designing the unrolled network for a robust GSD problem,
which will have significant positive societal impact. There are no
evidently negative societal impact as far as we know.

\section{Proofs and More Discussions}

\subsection{Proof of Proposition \ref{thm:SGC}}
\begin{proof}
With $\alpha=0$ and $r(\mathbf{H})=\beta\bigl\Vert\mathbf{H}\bigr\Vert_{\mathbf{I}-\mathbf{T}_{\beta}}^{2}$,
the GSD problem \eqref{Formulation: GSD reg} becomes
\[
\begin{aligned} & \underset{\mathbf{H}}{\mathrm{min}} &  & \beta\bigl\Vert\mathbf{B}\mathbf{H}\bigr\Vert_{\mathbf{T}_{\beta}}^{2}+\beta\bigl\Vert\mathbf{H}\bigr\Vert_{\mathbf{I}-\mathbf{T}_{\beta}}^{2}.\end{aligned}
\]
For the above problem, the $k$-th ($k=1,\ldots,K$) unrolled GD step
in \eqref{Unrolling Step: GD} accordingly turns to be
\begin{align*}
\mathbf{H}^{(k)} & =\mathbf{H}^{(k-1)}-\eta^{(k)}\left(2\beta^{(k)}\mathbf{B}^{T}\mathbf{B}\mathbf{H}^{(k-1)}\mathbf{T}_{\beta}^{(k)}+2\beta^{(k)}\mathbf{H}^{(k-1)}\bigl(\mathbf{I}-\mathbf{T}_{\beta}^{(k)}\bigr)\right)\\
 & =\mathbf{H}^{(k-1)}-\eta^{(k)}\left(2\beta^{(k)}(\mathbf{I}-\mathbf{A})\mathbf{H}^{(k-1)}\mathbf{T}_{\beta}^{(k)}+2\beta^{(k)}\mathbf{H}^{(k-1)}\bigl(\mathbf{I}-\mathbf{T}_{\beta}^{(k)}\bigr)\right)\\
 & =\mathbf{H}^{(k-1)}\left(\mathbf{I}-2\eta^{(k)}\beta^{(k)}\mathbf{T}_{\beta}^{(k)}-2\eta^{(k)}\beta^{(k)}\bigl(\mathbf{I}-\mathbf{T}_{\beta}^{(k)}\bigr)\right)+2\eta^{(k)}\beta^{(k)}\mathbf{A}\mathbf{H}^{(k-1)}\mathbf{T}_{\beta}^{(k)},
\end{align*}
where in the second line we have used the relation $\mathbf{I}-\mathbf{A}=\mathbf{B}^{T}\mathbf{B}$.
By choosing $\eta^{(k)}=\frac{1}{2\beta^{(k)}}$, the unrolled GD
step becomes
\[
\mathbf{H}^{(k)}=\mathbf{A}\mathbf{H}^{(k-1)}\mathbf{T}_{\beta}^{(k)}.
\]
Then, after stacking $K$ unrolled GD steps, we have 
\[
\bar{\mathbf{X}}=\mathbf{H}^{(K)}=\mathbf{A}^{K}\mathbf{H}^{(0)}\prod_{i=1}^{L}\mathbf{T}_{\beta}^{(k)}.
\]
By setting the initial point $\mathbf{H}^{(0)}=\mathbf{X}$ and reparameterizing
the weight $\prod_{i=1}^{L}\mathbf{T}_{\beta}^{(k)}=\mathbf{W}$,
we obtain the propagation mechanism in SGC, which completes the proof.
\end{proof}
In the above proof, we have assumed that the learnable weight matrix
$\mathbf{W}$ is induced from the model parameter $\boldsymbol{\Theta}$
in the GSD problem. However, another interesting observation is that
it can also be explained into $p_{\mathsf{pre}}$ or $p_{\mathsf{pos}}$.
Consider the GSD problem \eqref{Formulation: GSD reg} with $\alpha=0$,
$\mathbf{T}_{\alpha}=\mathbf{T}_{\beta}=\mathbf{I}$, and $r(\mathbf{H})=0$.
By choosing $\eta^{(k)}=\frac{1}{2\beta^{(k)}}$, the $k$-th ($k=1,\ldots,K$)
unrolled GD step is correspondingly given by
\[
\mathbf{H}^{(k)}=\mathbf{H}^{(k-1)}-2\eta^{(k)}\beta^{(k)}\mathbf{B}^{T}\mathbf{B}\mathbf{H}^{(k-1)}=\mathbf{A}\mathbf{H}^{(k-1)}.
\]
Stacking $K$ unrolled GD steps, we get the aggregation mechanism
as
\[
\mathbf{H}^{(K)}=\mathbf{A}^{K}\mathbf{H}^{(0)}.
\]
Setting $\mathbf{H}^{(0)}=\mathbf{X}$ and introducing the transformation
matrix $\mathbf{W}$ as a part of $P_{\mathsf{pre}}$ or $P_{\mathsf{pos}}$
lead to the SGC model. 

\subsection{More Discussions on SGC}

\textbf{Global smoothing properties of the SGC model.} From our analysis,
since SGC corresponds to a GSD problem with $\alpha=0$ (i.e., no
signal fidelity term), it is easy to conclude that the model only
performs global smoothing based on the graph topology. Coincidentally,
this property has been reported both numerically and theoretically
in some existing works \cite{li2018deeper,hoang2021revisiting}.

\textbf{Unique limit distributions of the SGC model.} It has been
discussed in \cite{xu2018representation,klicpera2019predict} that
under the assumption that the graph is irreducible and aperiodic,
the node representations of an infinite layer SGC model will converge
to a limit distribution corresponds to the random walk, which only
depends on the graph. While based on the GSD problem with the algorithm
unrolling perspective, different initial point $\mathbf{H}^{(0)}$
will lead to the same limit result as the GSD problem corresponding
to SGC is convex, which exactly coincides with and further corroborates
the analyses in \cite{xu2018representation,klicpera2019predict}.

\subsection{Proof of Proposition \ref{thm:APPNP}}
\begin{proof}
With $\mathbf{T}_{\alpha}=\mathbf{T}_{\beta}=\mathbf{I}$ and $r(\mathbf{H})=0$,
the GSD problem becomes the vanilla GSD problem, which has an analytical
solution \cite{stankovic2019introduction}. Since $\beta=1-\alpha$,
the analytical solution can be obtained by setting the gradient to
be zero, i.e.,
\[
2\alpha\left(\bar{\mathbf{X}}-\mathbf{X}\right)+2\left(1-\alpha\right)\mathbf{B}^{T}\mathbf{B}\bar{\mathbf{X}}=\mathbf{0},
\]
which gives
\[
\bar{\mathbf{X}}-\left(1-\alpha\right)\mathbf{A}\bar{\mathbf{X}}=\alpha\mathbf{X},
\]
or, equivalently,
\[
\bar{\mathbf{X}}=\alpha\left(\mathbf{I}-\left(1-\alpha\right)\mathbf{A}\right)^{-1}\mathbf{X},
\]
leading to the exact propagation scheme in PPNP model. Note that PPNP
was originally proposed based on the idea of personalized PageRank.
Here, we gave another equivalent interpretation based on GSD.

With $\beta=1-\alpha$, $\mathbf{T}_{\alpha}=\mathbf{T}_{\beta}=\mathbf{I}$,
and $r(\mathbf{H})=0$, the unrolled GD layer \eqref{Unrolling Step: GD}
for GSD is
\begin{equation}
\begin{aligned}\mathbf{H}^{(k)} & =\mathbf{H}^{(k-1)}-\eta^{(k)}\left(2\alpha^{(k)}\bigl(\mathbf{H}-\mathbf{X}\bigr)+2\bigl(1-\alpha^{(k)}\bigr)\mathbf{B}^{T}\mathbf{B}\mathbf{H}^{(k-1)}\right)\\
 & =\mathbf{H}^{(k-1)}-\eta^{(k)}\left(2\alpha^{(k)}\bigl(\mathbf{H}-\mathbf{X}\bigr)+2\bigl(1-\alpha^{(k)}\bigr)\bigl(\mathbf{I}-\mathbf{A}\bigr)\mathbf{H}^{(k-1)}\right)\\
 & =\mathbf{H}^{(k-1)}\left(1-2\eta^{(k)}\alpha^{(k)}-2\eta^{(k)}\bigl(1-\alpha^{(k)}\bigr)\right)+2\eta^{(k)}\bigl(1-\alpha^{(k)}\bigr)\mathbf{A}\mathbf{H}^{(k-1)}+2\eta^{(k)}\alpha^{(k)}\mathbf{X}.
\end{aligned}
\label{APPNP:unroll}
\end{equation}
With $\eta^{(k)}=\frac{1}{2}$ and $\alpha^{(k)}=\gamma$, $k=1,\ldots,K$,
the $k$-th unrolled GD layer becomes
\[
\mathbf{H}^{(k)}=\bigl(1-\gamma\bigr)\mathbf{A}\mathbf{H}^{(k-1)}+\gamma\mathbf{X},
\]
which is exactly the same with the $k$-th propagation layer in APPNP.
Note that the PPNP model can be naturally seen as applying infinitely
many unrolled GD layers.
\end{proof}

\subsection{Proof of Proposition \ref{thm: JKNet}}
\begin{proof}
With $\alpha=\beta$ and $\mathbf{T}_{\alpha}=\mathbf{I}-\mathbf{T}_{\beta}$,
the $k$-th ($k=1,\ldots,K$) unrolled GD layer \eqref{Unrolling Step: GD}
becomes
\begin{align*}
\mathbf{H}^{(k)} & =\mathbf{H}^{(k-1)}\left(\mathbf{I}-2\eta^{(k)}\alpha^{(k)}\mathbf{T}_{\alpha}^{(k)}-2\eta^{(k)}\alpha^{(k)}\bigl(\mathbf{I}-\mathbf{T}_{\alpha}^{(k)}\bigr)\right)\\
 & \hspace{1.8cm}+2\eta^{(k)}\alpha^{(k)}\mathbf{A}\mathbf{H}^{(k-1)}\bigl(\mathbf{I}-\mathbf{T}_{\alpha}^{(k)}\bigr)+2\eta^{(k)}\alpha^{(k)}\mathbf{X}\mathbf{T}_{\alpha}^{(k)}.
\end{align*}
By choosing $\eta^{(k)}=\frac{1}{2\alpha^{(k)}}$, we have
\[
\mathbf{H}^{(k)}=\mathbf{A}\mathbf{H}^{(k-1)}\bigl(\mathbf{I}-\mathbf{T}_{\alpha}^{(k)}\bigr)+\mathbf{X}\mathbf{T}_{\alpha}^{(k)}.
\]
Stacking $K$ unrolling GD layers, we have 
\[
\begin{aligned}\mathbf{H}^{(K)}= & \mathbf{A}\mathbf{H}^{(K-1)}\bigl(\mathbf{I}-\mathbf{T}_{\alpha}^{(K)}\bigr)+\mathbf{X}\mathbf{T}_{\alpha}^{(K)}\\
= & \mathbf{A}\left(\mathbf{A}\mathbf{H}^{(K-2)}\bigl(\mathbf{I}-\mathbf{T}_{\alpha}^{(K-1)}\bigr)+\mathbf{X}\mathbf{T}_{\alpha}^{(K-1)}\right)\bigl(\mathbf{I}-\mathbf{T}_{\alpha}^{(K)}\bigr)+\mathbf{X}\mathbf{T}_{\alpha}^{(K)}\\
= & \mathbf{A}^{2}\mathbf{H}^{(K-2)}\bigl(\mathbf{I}-\mathbf{T}_{\alpha}^{(K-1)}\bigr)\bigl(\mathbf{I}-\mathbf{T}_{\alpha}^{(K)}\bigr)+\mathbf{A}\mathbf{X}\mathbf{T}_{\alpha}^{(K-1)}\bigl(\mathbf{I}-\mathbf{T}_{\alpha}^{(K)}\bigr)+\mathbf{X}\mathbf{T}_{\alpha}^{(K)}\\
= & \ldots\\
= & \mathbf{A}^{K}\mathbf{H}^{(0)}\prod_{k=1}^{K}\bigl(\mathbf{I}-\mathbf{T}_{\alpha}^{(k)}\bigr)+\sum_{i=1}^{K}\mathbf{A}^{K-i}\mathbf{X}\mathbf{T}_{\alpha}^{(K-i)}\prod_{j=K-i+1}^{K}\bigl(\mathbf{I}-\mathbf{T}_{\alpha}^{(j)}\bigr).
\end{aligned}
\]
By reparameterizing $\prod_{k=1}^{K}\bigl(\mathbf{I}-\mathbf{T}_{\alpha}^{(k)}\bigr)=\mathbf{W}^{(K)}$
and $\mathbf{T}_{\alpha}^{(K-i)}\prod_{j=K-i+1}^{K}\bigl(\mathbf{I}-\mathbf{T}_{\alpha}^{(j)}\bigr)=\mathbf{W}^{(K-i)}$
for $i=1,\ldots,K$, and setting the initial point $\mathbf{H}^{(0)}=\mathbf{X}$,
we obtain the propagation mechanism in the JKNet model. 
\end{proof}

\subsection{Proof of Proposition \ref{thm: GPRGNN}}
\begin{proof}
With\textit{ $\beta=1-\alpha$, $\mathbf{T}_{\alpha}=\mathbf{T}_{\beta}=\mathbf{I}$,
and $r(\mathbf{H})=0$}, the GPRGNN corresponds to the same GSD problem
as APPNP and hence the $k$-th ($k=1,\ldots,K$) unrolled GD layer
is the same as \eqref{APPNP:unroll}. By setting $\eta^{(k)}=\frac{1}{2}$,
we have
\[
\mathbf{H}^{(k)}=\bigl(1-\alpha^{(k)}\bigr)\mathbf{A}\mathbf{H}^{(k-1)}+\alpha^{(k)}\mathbf{X}.
\]
Concatenating $K$ unrolled GD layers, we obtain
\begin{align*}
\mathbf{H}^{(K)} & =\bigl(1-\alpha^{(K)}\bigr)\mathbf{A}\mathbf{H}^{(K-1)}+\alpha^{(K)}\mathbf{X}\\
 & =\bigl(1-\alpha^{(K)}\bigr)\mathbf{A}\left(\bigl(1-\alpha^{(K-1)}\bigr)\mathbf{A}\mathbf{H}^{(K-2)}+\alpha^{(K-1)}\mathbf{X}\right)+\alpha^{(K)}\mathbf{X}\\
 & =\bigl(1-\alpha^{(K)}\bigr)\bigl(1-\alpha^{(K-1)}\bigr)\mathbf{A}^{2}\mathbf{H}^{(K-2)}+\bigl(1-\alpha^{(K)}\bigr)\alpha^{(K-1)}\mathbf{A}\mathbf{X}+\alpha^{(K)}\mathbf{X}\\
 & =\cdots\\
 & =\prod_{k=1}^{K}\bigl(1-\alpha^{(k)}\bigr)\mathbf{A}^{K}\mathbf{H}^{(0)}+\sum_{i=1}^{K}\alpha^{(K-i)}\Biggl(\prod_{j=K-i+1}^{K}\bigl(1-\alpha^{(j)}\bigr)\Biggr)\mathbf{A}^{K-i}\mathbf{X}.
\end{align*}
By reparameterizing $\prod_{k=1}^{K}\bigl(1-\alpha^{(k)}\bigr)=\gamma^{(K)}$
and $\alpha^{(K-i)}\prod_{j=K-i+1}^{K}\bigl(1-\alpha^{(j)}\bigr)=\gamma^{(K-i)}$
for $i=1,\ldots,K$, and setting the initial point $\mathbf{H}^{(0)}=\mathbf{X}$,
we get the propagation scheme in the GPRGNN model. 
\end{proof}

\subsection{Proof of Proposition \ref{thm:GCN}}
\begin{proof}
With $\alpha=0$ and $r(\mathbf{H})=\beta\bigl\Vert\mathbf{H}\bigr\Vert_{\mathbf{I}-\mathbf{T}_{\beta}}^{2}+I_{\{h_{ij}\geq0,\ \forall i,j\}}(\mathbf{H})$,
the $k$-th ($k=1,\ldots,L$) unrolled ProxGD layer \eqref{Unrolling Step: PGD}
becomes
\begin{align*}
\mathbf{H}^{(k)} & =\mathrm{ReLU}\biggl(\mathbf{H}^{(k-1)}-\eta^{(k)}\Bigl(2\beta^{(k)}\mathbf{B}^{T}\mathbf{B}\mathbf{H}^{(k-1)}\mathbf{T}_{\beta}^{(k)}+2\beta^{(k)}\mathbf{H}^{(k-1)}\bigl(\mathbf{I}-\mathbf{T}_{\beta}^{(k)}\bigr)\Bigr)\biggr)\\
 & =\mathrm{ReLU}\biggl(\mathbf{H}^{(k-1)}\Bigl(\mathbf{I}-2\eta^{(k)}\beta^{(k)}\mathbf{T}_{\beta}^{(k)}-2\eta^{(k)}\beta^{(k)}\bigl(\mathbf{I}-\mathbf{T}_{\beta}^{(k)}\bigr)\Bigr)+2\eta^{(k)}\beta^{(k)}\mathbf{A}\mathbf{H}^{(k-1)}\mathbf{T}_{\beta}^{(k)}\biggr).
\end{align*}
Choosing $\eta^{(k)}=\frac{1}{2\beta^{(k)}}$, we obtain the following
feature propagation process
\[
\mathbf{H}^{(k)}=\mathrm{ReLU}\bigl(\mathbf{A}\mathbf{H}^{(k-1)}\mathbf{T}_{\beta}^{(k)}\bigr).
\]
Setting $\mathbf{T}_{\beta}^{(k)}=\mathbf{W}^{(k)}$ recovers a $K$-layer
GCN model. 
\end{proof}

\subsection{More Discussions on GCN}

\textbf{Over-smoothing issues of the GCN model.} In the literature,
it was pointed out that the FA step in GCN, i.e., $\mathbf{A}\mathbf{H}^{(k-1)}$,
can be seen as a special form of Laplacian smoothing \cite{li2018deeper,yang2021rethinking}.
There were empirical evidence showing that the performance of GCNs
can severely degrade as the layer increases. With the above observations,
some papers suggested that the performance degradation in deep GCNs
is the consequence of the over-smoothing issue \cite{Model:PairNorm,cai2020note,hoang2021revisiting}.
Recently, contrary to the above argument some works argued that the
over-smoothing is just an artifact of theoretical analysis and is
not the main reason for performance degradation in deep GCNs \cite{yang2020revisiting,zhou2021understanding,cong2021provable}.
They pointed out that such discrepancy is because the previous over-simplified
analyses neglected the FT steps including the multiplication from
$\mathbf{W}^{(k)}$ and the nonlinear transformation $\mathrm{ReLU}$.
In this paper, based on the algorithm unrolling perspective, in which
we treat the consecutive propagation layers in GCN as a whole, we
can conclude that GCN cannot be seen as solely performing the goal
of signal smoothing due to the regularization term $r(\mathbf{H})$,
which supports the above claim that over-smoothing is not the main
reason for performance degradation in deep GCNs \cite{yang2020revisiting,zhou2021understanding,cong2021provable}.

\subsection{Proof of Proposition \ref{thm:GCNII}}
\begin{proof}
With $\beta=1-\alpha$, $\mathbf{T}_{\alpha}=\mathbf{T}_{\beta}$,
and $r(\mathbf{H})=\left\Vert \mathbf{H}\right\Vert _{\mathbf{I}-\mathbf{T}_{\beta}}^{2}+I_{\{h_{ij}\geq0,\ \forall i,j\}}(\mathbf{H})$,
the $k$-th ($k=1,\ldots,L$) unrolled ProxGD layer \eqref{Unrolling Step: PGD}
becomes
\begin{align*}
\mathbf{H}^{(k)} & =\mathrm{ReLU}\biggl(\mathbf{H}^{(k-1)}-\eta^{(k)}\Bigl(2\bigl(1-\beta^{(k)}\bigr)\bigl(\mathbf{H}-\mathbf{X}\bigr)\mathbf{T}_{\beta}^{(k)}+2\beta^{(k)}\mathbf{B}^{T}\mathbf{B}\mathbf{H}^{(k-1)}\mathbf{T}_{\beta}^{(k)}\\
 & \hspace{5cm}+2\mathbf{H}^{(k-1)}\bigl(\mathbf{I}-\mathbf{T}_{\beta}^{(k)}\bigr)\Bigr)\biggr)\\
 & =\mathrm{ReLU}\biggl(\mathbf{H}^{(k-1)}-\eta^{(k)}\Bigl(2\bigl(1-\beta^{(k)}\bigr)\bigl(\mathbf{H}-\mathbf{X}\bigr)\mathbf{T}_{\beta}^{(k)}+2\beta^{(k)}\bigl(\mathbf{I}-\mathbf{A}\bigr)\mathbf{H}^{(k-1)}\mathbf{T}_{\beta}^{(k)}\\
 & \hspace{5cm}+2\mathbf{H}^{(k-1)}\bigl(\mathbf{I}-\mathbf{T}_{\beta}^{(k)}\bigr)\Bigr)\biggr)\\
 & =\mathrm{ReLU}\biggl(\mathbf{H}^{(k-1)}\Bigl(\mathbf{I}-2\eta^{(k)}\bigl(1-\beta^{(k)}\bigr)\mathbf{T}_{\beta}^{(k)}-2\eta^{(k)}\beta^{(k)}\mathbf{T}_{\beta}^{(k)}-2\eta^{(k)}\bigl(\mathbf{I}-\mathbf{T}_{\beta}^{(k)}\bigr)\Bigr)\\
 & \hspace{5cm}+2\eta^{(k)}\beta^{(k)}\mathbf{A}\mathbf{H}^{(k-1)}\mathbf{T}_{\beta}^{(k)}+2\eta^{(k)}\bigl(1-\beta^{(k)}\bigr)\mathbf{X}\mathbf{T}_{\beta}^{(k)}\biggr).
\end{align*}
By choosing $\eta^{(k)}=\frac{1}{2}$, we have
\[
\mathbf{H}^{(k)}=\mathrm{ReLU}\Bigl(\beta^{(k)}\mathbf{A}\mathbf{H}^{(k-1)}\mathbf{T}_{\beta}^{(k)}+\bigl(1-\beta^{(k)}\bigr)\mathbf{X}\mathbf{T}_{\beta}^{(k)}\Bigr).
\]
Further setting $\beta^{(k)}=\zeta$, $k=1,\ldots,K$ and reparameterizing
$\mathbf{T}_{\beta}^{(k)}=\xi\mathbf{W}^{(k)}+\left(1-\xi\right)\mathbf{I}$,
we get the GCNII model. 
\end{proof}

\subsection{Proof of Proposition \ref{thm:AirGNN}}
\begin{proof}
Using the ProxGD algorithm with $\alpha^{(k)}=\frac{\gamma}{1-\gamma}\beta^{(k)}$
and stepsize $\eta^{(k)}=\frac{1}{2(1-\alpha^{(k)})}$, we obtain
the following iteration:
\begin{align*}
\mathbf{H}^{(k)} & =\text{arg}\min_{\mathbf{Y}}\ \bigl(1-\beta^{(k)}\bigr)\bigl\Vert\mathbf{Y}-\mathbf{X}\bigr\Vert_{2,1}+\frac{1}{2\eta^{(k)}}\bigl\Vert\mathbf{Y}-\bigl(\mathbf{H}^{(k-1)}-2\eta^{(k)}\beta^{(k)}(\mathbf{I}-\mathbf{A})\mathbf{H}^{(k-1)}\bigr)\bigr\Vert_{F}^{2}\\
 & =\text{arg}\min_{\mathbf{Y}}\ \bigl(1-\beta^{(k)}\bigr)\bigl\Vert\mathbf{Y}-\mathbf{X}\bigr\Vert_{2,1}\\
 & \hspace{4cm}+\frac{1}{2\eta^{(k)}}\bigl\Vert\mathbf{Y}-(1-2\eta^{(k)}\beta^{(k)})\mathbf{H}^{(k-1)}-2\eta^{(k)}\beta^{(k)}\mathbf{A}\mathbf{H}^{(k-1)}\bigr\Vert_{F}^{2}.
\end{align*}
By choosing $\eta^{(k)}=\frac{1}{2\beta^{(k)}}$, we have
\begin{align*}
\mathbf{H}^{(k)} & =\text{arg}\min_{\mathbf{Y}}\ \bigl(1-\beta^{(k)}\bigr)\bigl\Vert\mathbf{Y}-\mathbf{X}\bigr\Vert_{2,1}+\beta^{(k)}\bigl\Vert\mathbf{Y}-\mathbf{A}\mathbf{H}^{(k-1)}\bigr\Vert_{F}^{2}.
\end{align*}
The above proximal minimization problem can be computed analytically
where the $i$-th row of $\mathbf{H}^{(k)}$, i.e. $\mathbf{h}_{i}^{(k)}$,
is given as follows \cite{liu2021graph}: 
\[
\mathbf{h}_{i}^{(k)}=\mathrm{ReLU}\left(1-\frac{1-\beta^{(k)}}{2\beta^{(k)}||[\mathbf{A}\mathbf{H}^{(k-1)}]_{i}-\mathbf{x}_{i}||_{2}}\right)\bigl([\mathbf{A}\mathbf{H}^{(k-1)}]_{i}-\mathbf{x}_{i}\bigr)+\mathbf{x}_{i}.
\]
By reparameterizing $\beta^{(k)}=\gamma$ for $k=1,\ldots,K$, we
arrive at the AirGNN model. 
\end{proof}

\subsection{Proof of Proposition \ref{UGDGNN}}
\begin{proof}
By stacking $K$ unrolled GD layers \eqref{Unrolling Step: GD} with
initialization $\mathbf{H}^{(0)}=\mathbf{X}$, we have
\begin{align*}
\mathbf{H}^{(K)}= & \mathbf{H}^{(K-1)}\bigl(\mathbf{I}-2\eta^{(K)}\alpha^{(K)}\mathbf{T}_{\alpha}^{(K)}-2\eta^{(K)}\beta^{(K)}\mathbf{T}_{\beta}^{(K)}\bigr)\\
 & \hspace{5.5cm}+2\eta^{(K)}\beta^{(K)}\mathbf{A}\mathbf{H}^{(K-1)}\mathbf{T}_{\beta}^{(K)}+2\eta^{(K)}\alpha^{(K)}\mathbf{X}\mathbf{T}_{\alpha}^{(K)}\\
= & \Bigl(\mathbf{H}^{(K-2)}\bigl(\mathbf{I}-2\eta^{(K-1)}\alpha^{(K-1)}\mathbf{T}_{\alpha}^{(K-1)}-2\eta^{(K-1)}\beta^{(K-1)}\mathbf{T}_{\beta}^{(K-1)}\bigr)\\
 & \hspace{0.9cm}+2\eta^{(K-1)}\beta^{(K-1)}\mathbf{A}\mathbf{H}^{(K-2)}\mathbf{T}_{\beta}^{(K-1)}+2\eta^{(K-1)}\alpha^{(K-1)}\mathbf{X}\mathbf{T}_{\alpha}^{(K-1)}\Bigr)\\
 & \hspace{0.9cm}\times\bigl(\mathbf{I}-2\eta^{(K)}\alpha^{(K)}\mathbf{T}_{\alpha}^{(K)}-2\eta^{(K)}\beta^{(K)}\mathbf{T}_{\beta}^{(K)}\bigr)+2\eta^{(K)}\alpha^{(K)}\mathbf{X}\mathbf{T}_{\alpha}^{(K)}\\
 & \hspace{0.9cm}+2\eta^{(K)}\beta^{(K)}\mathbf{A}\Bigl(\mathbf{H}^{(K-2)}\bigl(\mathbf{I}-2\eta^{(K-1)}\alpha^{(K-1)}\mathbf{T}_{\alpha}^{(K-1)}-2\eta^{(K-1)}\beta^{(K-1)}\mathbf{T}_{\beta}^{(K-1)}\bigr)\\
 & \hspace{0.9cm}+2\eta^{(K-1)}\beta^{(K-1)}\mathbf{A}\mathbf{H}^{(K-2)}\mathbf{T}_{\beta}^{(K-1)}+2\eta^{(K-1)}\alpha^{(K-1)}\mathbf{X}\mathbf{T}_{\alpha}^{(K-1)}\Bigr)\mathbf{T}_{\beta}^{(K)}\\
= & 2\eta^{(K)}\beta^{(K)}2\eta^{(K-1)}\beta^{(K-1)}\mathbf{A}^{2}\mathbf{H}^{(K-2)}\mathbf{T}_{\beta}^{(K-1)}\mathbf{T}_{\beta}^{(K)}\\
 & \hspace{0.9cm}+\mathbf{A}\mathbf{H}^{(K-2)}\Bigl(2\eta^{(K-1)}\beta^{(K-1)}\mathbf{T}_{\beta}^{(K-1)}\bigl(\mathbf{I}-2\eta^{(K)}\alpha^{(K)}\mathbf{T}_{\alpha}^{(K)}-2\eta^{(K)}\beta^{(K)}\mathbf{T}_{\beta}^{(K)}\bigr)\\
 & \hspace{0.9cm}+2\eta^{(K)}\beta^{(K)}\bigl(\mathbf{I}-2\eta^{(K-1)}\alpha^{(K-1)}\mathbf{T}_{\alpha}^{(K-1)}-2\eta^{(K-1)}\beta^{(K-1)}\mathbf{T}_{\beta}^{(K-1)}\bigr)\mathbf{T}_{\beta}^{(K)}\Bigr)\\
 & \hspace{0.9cm}+\mathbf{H}^{(K-2)}\bigl(\mathbf{I}-2\eta^{(K-1)}\alpha^{(K-1)}\mathbf{T}_{\alpha}^{(K-1)}-2\eta^{(K-1)}\beta^{(K-1)}\mathbf{T}_{\beta}^{(K-1)}\bigr)\\
 & \hspace{0.9cm}\times\bigl(\mathbf{I}-2\eta^{(K)}\alpha^{(K)}\mathbf{T}_{\alpha}^{(K)}-2\eta^{(K)}\beta^{(K)}\mathbf{T}_{\beta}^{(K)}\bigr)\\
 & \hspace{0.9cm}+2\eta^{(K)}\beta^{(K)}2\eta^{(K-1)}\alpha^{(K-1)}\mathbf{A}\mathbf{X}\mathbf{T}_{\alpha}^{(K-1)}\mathbf{T}_{\beta}^{(K)}+\mathbf{X}\Bigl(2\eta^{(K)}\alpha^{(K)}\mathbf{T}_{\alpha}^{(K)}\\
 & \hspace{0.9cm}+2\eta^{(K-1)}\alpha^{(K-1)}\mathbf{T}_{\alpha}^{(K-1)}\bigl(\mathbf{I}-2\eta^{(K)}\alpha^{(K)}\mathbf{T}_{\alpha}^{(K)}-2\eta^{(K)}\beta^{(K)}\mathbf{T}_{\beta}^{(K)}\bigr)\Bigr)\\
= & \mathbf{A}^{K}\mathbf{X}\prod_{k=1}^{K}2\eta^{(k)}\beta^{(k)}\mathbf{T}_{\beta}^{(k)}\\
 & +\cdots\\
 & +\mathbf{A}^{0}\mathbf{X}\biggl(2\eta^{(K)}\alpha^{(K)}\mathbf{T}_{\alpha}^{(K)}+\Bigl(2\eta^{(K-1)}\alpha^{(K-1)}\mathbf{T}_{\alpha}^{(K-1)}+\bigl(2\eta^{(K-2)}\alpha^{(K-2)}\mathbf{T}_{\alpha}^{(K-2)}+\cdots\bigr)\\
 & \times\bigl(\mathbf{I}-2\eta^{(K-1)}\alpha^{(K-1)}\mathbf{T}_{\alpha}^{(K-1)}-2\eta^{(K-1)}\beta^{(K-1)}\mathbf{T}_{\beta}^{(K-1)}\bigr)\Bigr)\\
 & \times\bigl(\mathbf{I}-2\eta^{(K)}\alpha^{(K)}\mathbf{T}_{\alpha}^{(K)}-2\eta^{(K)}\beta^{(K)}\mathbf{T}_{\beta}^{(K)}\bigr)\biggr).
\end{align*}
To reduce the complexity of the unrolled network, we reparameterize
it into the following form
\[
\bar{\mathbf{X}}=\sum_{k=0}^{K}\gamma^{(k)}\mathbf{A}^{k}\mathbf{X}\left(\zeta^{(k)}\mathbf{I}+\xi^{(k)}\mathbf{W}^{(k)}\right)
\]
with $\gamma^{(k)}$, $\zeta^{(k)}$, $\xi^{(k)}$, and $\mathbf{W}^{(k)}$
being the learnable model parameters. Note that we have left the identity
matrix outside of the weight matrix to promote the optimal weight
matrices with small norms and reduce the model complexity, which guarantees
better generalization ability \cite{cong2021provable} and has been
shown to be useful particularly in semi-supervised learning tasks
\cite{chen2020simple}. In the experiments, we further set $\xi^{(k)}=1-\zeta^{(k)}$
to reduce the parameter space. Note that such a further restriction
does not change the expressive power of UGDGNN. Similarly, one can
also choose to reduce the parameter $\gamma^{(k)}$ by merging it
into $\zeta^{(k)}$ and $\xi^{(k)}$.
\end{proof}

\subsection{Proof of Theorem \ref{thm: Spectral} (Expressive Power Analysis
of UGDGNN in the Spectral Domain)\label{subsec:Proof-of-Theorem}}
\begin{proof}
The propagation scheme of a $K$ layer UGDGNN is
\begin{align}
\bar{\mathbf{X}} & =\sum_{k=0}^{K}\gamma^{(k)}\mathbf{A}^{k}\mathbf{X}\bigl(\zeta^{(k)}\mathbf{I}+\xi^{(k)}\mathbf{W}^{(k)}\bigr)\nonumber \\
 & =\sum_{k=0}^{K}\gamma^{(k)}\bigl(\mathbf{I}-\mathbf{L}\bigr)^{k}\mathbf{X}\bigl(\zeta^{(k)}\mathbf{I}+\xi^{(k)}\mathbf{W}^{(k)}\bigr).\label{eq:UGDGNN}
\end{align}
On the other hand, an order $K$ polynomial frequency filter \cite{shuman2013emerging}
can be expressed as
\begin{align}
\left(\sum_{k=0}^{K}\theta_{k}\mathbf{L}^{k}\right)\mathbf{X} & =\left(\sum_{k=0}^{K}\theta_{k}\left(\mathbf{I}-\left(\mathbf{I}-\mathbf{L}\right)\right)^{k}\right)\mathbf{X}\nonumber \\
 & =\left(\sum_{k=0}^{K}\theta_{k}\left(\sum_{i=0}^{k}\left(-1\right)^{i}\left(\begin{array}{c}
k\\
i
\end{array}\right)\left(\mathbf{I}-\mathbf{L}\right)^{i}\right)\right)\mathbf{X}\nonumber \\
 & =\sum_{k=0}^{K}\sum_{i=0}^{k}\theta_{k}\left(-1\right)^{i}\left(\begin{array}{c}
k\\
i
\end{array}\right)\left(\mathbf{I}-\mathbf{L}\right)^{i}\mathbf{X}\nonumber \\
 & =\sum_{i=0}^{K}\left(\sum_{k=i}^{K}\theta_{k}\left(-1\right)^{i}\left(\begin{array}{c}
k\\
i
\end{array}\right)\right)\left(\mathbf{I}-\mathbf{L}\right)^{i}\mathbf{X},\label{eq:filter}
\end{align}
where $\left(\begin{array}{c}
k\\
i
\end{array}\right)$ represents the number of combinations for ``$k$ choose $i$'' and
in the last line we have exchanged the order of the double summations. 

Comparing \eqref{eq:UGDGNN} and \eqref{eq:filter}, the UGDGNN model
with $\gamma^{(i)}=\sum_{k=i}^{K}\theta_{k}\left(-1\right)^{i}\left(\begin{array}{c}
k\\
i
\end{array}\right)$, $\zeta^{(i)}=1$, and $\xi^{(i)}=0$ for $i=1,\ldots,K$ is equivalent
to a polynomial frequency filter, which completes the proof.
\end{proof}

\section{Discussions on the Expressive Power of UGDGNN in the Vertex Domain}

Since the UGDGNN model is derived based on concatenating the general
unrolled GD layers \eqref{Unrolling Step: GD}, it is expected to
be more expressive than other GNN models explained via unrolled GD
layers with specific parameters. In the following, we give the connections
between UGDGNN and several existing GNNs that can be seen as unrolled
GD networks.
\begin{itemize}
\item The SGC model can be seen as a UGDGNN with $\gamma^{(K)}=1$, $\zeta^{(K)}=0$,
$\xi^{(K)}=1$, and $\gamma^{(k)}=0$ for $k=1,\ldots,K-1$. 
\item The APPNP model can be seen as a UGDGNN with $\zeta^{(k)}=1$ and
$\xi^{(k)}=0$, $k=1,\ldots,K$, while $\gamma^{(K)}=\left(1-\alpha\right)^{K}$
and $\gamma^{(k)}=\alpha\sum_{k=0}^{K-1}\left(1-\alpha\right)^{k}$
for $k=1,\ldots,K-1$.
\item The JKNet model can be seen as a UGDGNN with $\gamma^{(k)}=1$, $\zeta^{(k)}=0$,
and $\xi^{(k)}=1$ for $k=1,\ldots,K$.
\item The GPRGNN model can be seen as a UGDGNN with $\zeta^{(k)}=1$ and
$\xi^{(k)}=0$ for $k=1,\ldots,K$.
\end{itemize}

\section{Expressive Power Analysis of Existing GNNs in the Spectral Domain
\label{Appendix: Spectral}}

\subsection{Expressive Power of SGC and GCN}

It has been shown that a $K$-layer SGC act as a frequency filter
of order $K$ with fixed coefficients \cite{wu2019simplifying,klicpera2019predict,chen2020simple}.
Recalling that a $K$-layer SGC is
\[
\bar{\mathbf{X}}=\mathbf{A}^{K}\mathbf{X}\mathbf{W}=\left(\mathbf{I}-\mathbf{L}\right)^{K}\mathbf{X}\mathbf{W}=\sum_{k=0}^{K}\left(-1\right)^{k}\left(\begin{array}{c}
K\\
k
\end{array}\right)\mathbf{L}^{k}\mathbf{X}\mathbf{W}.
\]
It can be seen as an order-$K$ frequency filter with fixed coefficients
$\theta_{k}=\left(-1\right)^{k}\left(\begin{array}{c}
K\\
k
\end{array}\right)$, $k=1,\ldots,K$ and $\mathbf{W}=\mathbf{I}$. 

If we neglect the activation functions in GCN or, equivalently, assume
all the intermediate node representations are positive and set all
the weight matrices to be identity matrices, the GCN model becomes
the SGC model, and therefore it also act as an order-$K$ frequency
filter with fixed coefficients. Such a frequency filter with fixed
coefficients inevitably has limited expressive power.

\subsection{Expressive Power of APPNP}

The propagation scheme of a $K$-layer APPNP is
\[
\bar{\mathbf{X}}=\mathbf{H}^{(K)},\ \ \mathbf{H}^{(k)}=\left(1-\alpha\right)\mathbf{A}\mathbf{H}^{(k-1)}+\alpha\mathbf{X},\ \ k=1,\ldots,K.
\]
With $\mathbf{H}^{(0)}=\mathbf{X}$, by stacking the $K$ layers together,
we have \cite{klicpera2019predict}
\[
\bar{\mathbf{X}}=\left(\left(1-\alpha\right)^{K}\mathbf{A}^{K}+\alpha\sum_{k=0}^{K-1}\left(1-\alpha\right)^{k}\mathbf{A}^{k}\right)\mathbf{X}.
\]
Suppose $K$ is sufficiently large, we have $\left(1-\alpha\right)^{K}\rightarrow0$,
leading to
\begin{align*}
\bar{\mathbf{X}} & =\alpha\sum_{k=0}^{K-1}\left(1-\alpha\right)^{k}\mathbf{A}^{k}\mathbf{X}\\
 & =\alpha\sum_{k=0}^{K-1}\left(1-\alpha\right)^{k}\left(\mathbf{I}-\mathbf{L}\right)^{k}\mathbf{X}\\
 & =\alpha\sum_{k=0}^{K-1}\left(1-\alpha\right)^{k}\sum_{i=0}^{k}\left(\begin{array}{c}
k\\
i
\end{array}\right)\left(-1\right)^{i}\mathbf{L}^{i}\mathbf{X}\\
 & =\sum_{k=0}^{K-1}\sum_{i=0}^{k}\alpha\left(1-\alpha\right)^{k}\left(\begin{array}{c}
k\\
i
\end{array}\right)\left(-1\right)^{i}\mathbf{L}^{i}\mathbf{X}\\
 & =\sum_{i=0}^{K-1}\sum_{k=i}^{K-1}\alpha\left(1-\alpha\right)^{k}\left(\begin{array}{c}
k\\
i
\end{array}\right)\left(-1\right)^{i}\mathbf{L}^{i}\mathbf{X}.
\end{align*}
Therefore, we can conclude that the APPNP model act as an order-$K$
frequency filter with fixed coefficients $\theta_{k}=\sum_{k=k}^{K-1}\alpha\left(1-\alpha\right)^{k}\left(\begin{array}{c}
k\\
k
\end{array}\right)\left(-1\right)^{k}$, $k=1,\ldots,K$, and $\theta_{K}=0$.

\subsection{Expressive Power of GCNII}

Recalling that the propagation scheme in GCNII is
\[
\bar{\mathbf{X}}=\mathbf{H}^{(K)},\ \ \mathbf{H}^{(k)}=\mathrm{ReLU}\Bigl(\bigl(\beta\mathbf{A}\mathbf{H}^{(k-1)}+(1-\beta)\mathbf{X}\bigr)\bigl(\gamma\mathbf{W}^{(k)}+(1-\gamma)\mathbf{I}\bigr)\Bigr),\ \ k=1,\ldots,K.
\]
Similar to the analysis for GCN, we neglect the activation functions.
Assuming $\beta=\frac{1}{2}$ and $\gamma\mathbf{W}^{(k)}+(1-\gamma)\mathbf{I}=2w^{(k)}\mathbf{I}$,
we have
\begin{align*}
\mathbf{H}^{(k)} & =\bigl(\mathbf{A}\mathbf{H}^{(k-1)}+\mathbf{X}\bigr)w^{(k)}\mathbf{I}\\
 & =w^{(k)}\bigl(\mathbf{I}-\mathbf{L}\bigr)\mathbf{H}^{(k-1)}+w^{(k)}\mathbf{X}.
\end{align*}
Stacking $K$ layers and setting $\mathbf{H}^{(0)}=\mathbf{X}$, we
obtain
\[
\bar{\mathbf{X}}=\sum_{k=0}^{K}\left(\prod_{k=K-k}^{K}w^{(k)}\right)\left(\mathbf{I}-\mathbf{L}\right)^{k}\mathbf{X}.
\]
As shown in Appendix \ref{Appendix: Spectral}, a frequency filter
of order-$K$ can be expressed as
\[
\left(\sum_{k=0}^{K}\theta_{k}\mathbf{L}^{k}\right)\mathbf{X}=\sum_{i=0}^{K}\left(\sum_{k=i}^{K}\theta_{k}\left(-1\right)^{i}\left(\begin{array}{c}
k\\
i
\end{array}\right)\right)\left(\mathbf{I}-\mathbf{L}\right)^{i}\mathbf{X}.
\]
Comparing the two equations above, we can find a GCNII model can express
an arbitrary frequency filter with nonzero coefficients with \cite{chen2020simple}
\[
w^{(k)}=\frac{\sum_{k=i}^{K}\theta_{k}\left(-1\right)^{i}\left(\begin{array}{c}
k\\
i
\end{array}\right)}{\sum_{k=i-1}^{K}\theta_{k}\left(-1\right)^{i-1}\left(\begin{array}{c}
k\\
i-1
\end{array}\right)}.
\]

As for the JKNet model and the GPRGNN model, they have the same spectral
expressive power as UGDGNN and can be proved in a fashion similar
as in Appendix \ref{subsec:Proof-of-Theorem}.

From the above analyses, we can conclude that UGDGNN, JKNet, GPRGNN
has better spectral expressive power in comparison with SGC, GCN,
APPNP, and GCNII.

\section{Experimental Details \label{sec:Experiment-Details}}

\subsection{Datasets}

To ensure a fair comparison, we choose the same data splits as in
\cite{yang2016revisiting} for the Cora, CiteSeer, and PubMed datasets
\cite{sen2008collective}, i.e., we use 20 labeled nodes per class
as the training set, 500 nodes as the validation set, and 1000 nodes
as the test set. For co-authorship and co-purchase graph datasets
\cite{shchur2018pitfalls}, we use 20 labeled nodes per class as the
training set, 30 node per class as the validation set, and the rest
as the test set. For OGBN-ArXiv \cite{hu2020open}, we use the standard
leaderboard splitting, i.e. paper published until 2017 for training,
papers published in 2018 for validation, and papers published since
2019 for testing. The statistics for all the benchmark datasets are
summarized in Table \ref{tab:Statistics-of-datasets.}.

\begin{table}[h]
\centering{}\caption{Statistics of benchmark datasets.\label{tab:Statistics-of-datasets.}}
\resizebox{1 \textwidth}{!}{%
\begin{tabular}{cccccccc}
\hline 
Dataset & Nodes & Edges & Features & Classes & Training Nodes & Validation Nodes & Test Nodes\tabularnewline
\hline 
Cora & 2708 & 5429 & 1433 & 7 & 20 per class & 500 & 1000\tabularnewline
\hline 
Citeseer & 3327 & 4732 & 3703 & 6 & 20 per class & 500 & 1000\tabularnewline
\hline 
Pubmed & 19717 & 44338 & 500 & 3 & 20 per class & 500 & 1000\tabularnewline
\hline 
Coauthor CS & 18333 & 81894 & 6805 & 15 & 20 per class & 30 per class & Rest nodes\tabularnewline
\hline 
Coauthor Physics & 34493 & 247962 & 8415 & 5 & 20 per class & 30 per class & Rest nodes\tabularnewline
\hline 
Amazon Photo & 7487 & 119043 & 745 & 8 & 20 per class & 30 per class & Rest nodes\tabularnewline
\hline 
OGBN-ArXiv & 169343 & 1166243 & 128 & 40 & Papers until 2017 & Papers in 2018 & Paper since 2019\tabularnewline
\hline 
\end{tabular}}
\end{table}

\subsection{Implementation Details}

In this paper, experiments are conducted either on an 11G RTX 2080Ti
GPU or on a 48G RTX A6000 GPU. The implementation is based on PyTorch
Geometric \cite{Fey/Lenssen/2019} and the code is provided in the
supplemental materials. We train all the GNN models with a maximum
of 1000 training epochs and the Adam optimizer \cite{kingma2015adam}
is adopted for model training. The reported results are all averaged
values over ten times of independent repetitions.

To ensure fair comparisons, we fix the size of hidden layers uniformly
as 64 for all models and on all datasets. Other hyperparameters are
tuned with grid search based on the validation results. Actually,
we have observed that some models can achieve better results than
the original results reported. For all models, the learning rate is
chosen from \{0.005, 0.01, 0.05\} and the dropout rate is chosen from
\{0.1, 0.5, 0.8\}. We also apply the $\ell_{2}$ regularization technique
on the weight parameters for all models, where the weight decay parameter
is chosen from \{0, 5e-5, 5e-4\}. The layers for GCN and SGC are set
as 2, while the layers for other models are selected from \{5, 10,
20\}. The teleport probability in APPNP and GPRGNN (note that teleport
probability is used just for initialization in GPRGNN) is selected
from \{0.1, 0.2, 0.5, 0.9\}. The best sets of hyperparameters for
all models on all datasets are are summarized in Table \ref{tab:Hyperparametes}.
Note that the hyperparameters used for generating the results in Figure
1 are the same as those in Table 2 (from the main body of the paper).
That is, hyperparameters like learning rate, weight decay, and dropout
are not specifically tuned for different layers. Therefore, there
is a chance that the results reported in Figure 1 can be further improved
by carefully tuning the hyperparameters.

\begin{table}[h]
\begin{centering}
\caption{Hyperparameters (for Table 2 of the main paper) selected based on
the validation set. \label{tab:Hyperparametes}}
\par\end{centering}
\centering{}\resizebox{1 \textwidth}{!}{%
\begin{tabular}{c|c|cccccccc}
\hline 
\textbf{Dataset} & \textbf{Hyperparameter} & GCN & SGC & APPNP & JKNet & GCNII & DAGNN & GPRGNN & UGDGNN\tabularnewline
\hline 
\multirow{5}{*}{Cora} & learning rate & 0.01 & 0.05 & 0.005 & 0.05 & 0.01 & 0.01 & 0.01 & 0.005\tabularnewline
 & weight decay & 5e-4 & 5e-5 & 5e-5 & 5e-4 & 5e-4 & 5e-4 & 5e-4 & 5e-4\tabularnewline
 & dropout & 0.8 & {*} & 0.1 & 0.5 & 0.5 & 0.8 & 0.5 & 0.8\tabularnewline
 & layer & {*} & {*} & 5 & 5 & 20 & 10 & 10 & 5\tabularnewline
 & alpha & {*} & {*} & 0.1 & {*} & {*} & {*} & 0.1 & {*}\tabularnewline
\hline 
\multirow{5}{*}{CiteSeer} & learning rate & 0.05 & 0.05 & 0.01 & 0.05 & 0.05 & 0.005 & 0.005 & 0.01\tabularnewline
 & weight decay & 5e-4 & 5e-5 & 5e-4 & 5e-4 & 0 & 5e-4 & 5e-4 & 5e-5\tabularnewline
 & dropout & 0.5 & {*} & 0.1 & 0.8 & 0.5 & 0.5 & 0.5 & 0.8\tabularnewline
 & layer & {*} & {*} & 5 & 5 & 20 & 5 & 10 & 5\tabularnewline
 & alpha & {*} & {*} & 0.2 & {*} & {*} & {*} & 0.1 & {*}\tabularnewline
\hline 
\multirow{5}{*}{PubMed} & learning rate & 0.005 & 0.05 & 0.005 & 0.05 & 0.05 & 0.005 & 0.01 & 0.005\tabularnewline
 & weight decay & 5e-4 & 5e-5 & 5e-4 & 5e-4 & 0 & 5e-4 & 5e-4 & 5e-4\tabularnewline
 & dropout & 0.1 & {*} & 0.1 & 0.8 & 0.5 & 0.8 & 0.1 & 0.1\tabularnewline
 & layer & {*} & {*} & 5 & 5 & 20 & 20 & 20 & 5\tabularnewline
 & alpha & {*} & {*} & 0.2 & {*} & {*} & {*} & 0.5 & {*}\tabularnewline
\hline 
\multirow{5}{*}{CS} & learning rate & 0.005 & 0.005 & 0.05 & 0.05 & 0.01 & 0.005 & 0.01 & 0.005\tabularnewline
 & weight decay & 5e-5 & 5e-5 & 5e-5 & 5e-5 & 5e-4 & 5e-4 & 0 & 5e-5\tabularnewline
 & dropout & 0.8 & {*} & 0.8 & 0.8 & 0.1 & 0.8 & 0.5 & 0.1\tabularnewline
 & layer & {*} & {*} & 20 & 5 & 20 & 10 & 5 & 5\tabularnewline
 & alpha & {*} & {*} & 0.2 & {*} & {*} & {*} & 0.1 & {*}\tabularnewline
\hline 
\multirow{5}{*}{Physics} & learning rate & 0.01 & 0.005 & 0.005 & 0.01 & 0.05 & 0.005 & 0.005 & 0.005\tabularnewline
 & weight decay & 0 & 0 & 5e-4 & 0 & 5e-4 & 0 & 5e-5 & 0\tabularnewline
 & dropout & 0.5 & {*} & 0.1 & 0.8 & 0.5 & 0.8 & 0.1 & 0.8\tabularnewline
 & layer & {*} & {*} & 10 & 5 & 20 & 5 & 20 & 5\tabularnewline
 & alpha & {*} & {*} & 0.1 & {*} & {*} & {*} & 0.5 & {*}\tabularnewline
\hline 
\multirow{5}{*}{Photo} & learning rate & 0.01 & 0.05 & 0.05 & 0.05 & 0.01 & 0.05 & 0.005 & 0.005\tabularnewline
 & weight decay & 0 & 0 & 5e-5 & 5e-5 & 5e-5 & 5e-4 & 5e-4 & 0\tabularnewline
 & dropout & 0.8 & {*} & 0.1 & 0.5 & 0.1 & 0.5 & 0.1 & 0.5\tabularnewline
 & layer & {*} & {*} & 10 & 10 & 5 & 5 & 5 & 5\tabularnewline
 & alpha & {*} & {*} & 0.1 & {*} & {*} & {*} & 0.1 & {*}\tabularnewline
\hline 
\multirow{5}{*}{ArXiv} & learning rate & 0.005 & 0.05 & 0.01 & 0.005 & 0.005 & 0.005 & 0.01 & 0.005\tabularnewline
 & weight decay & 0 & 0 & 0 & 5e-4 & 5e-4 & 0 & 5e-4 & 0\tabularnewline
 & dropout & 0.1 & {*} & 0.1 & 0.1 & 0.1 & 0.1 & 0.1 & 0.1\tabularnewline
 & layer & {*} & {*} & 5 & 10 & 10 & 20 & 5 & 10\tabularnewline
 & alpha & {*} & {*} & 0.1 & {*} & {*} & {*} & 0.1 & {*}\tabularnewline
\hline 
\end{tabular}}
\end{table}
 
\end{document}